\documentclass[10pt,twocolumn,letterpaper]{article}

\usepackage{etoolbox}

\newbool{flagexemplars} 
\setbool{flagexemplars}{false}

\usepackage{makecell}

\usepackage{iccv_author_kit/iccv}              

\newcommand{\ts}{\textsuperscript}

\definecolor{cvprblue}{rgb}{0.21,0.49,0.74}
\usepackage[pagebackref,breaklinks,colorlinks,allcolors=cvprblue]{hyperref}

\def\paperID{14385} 
\def\confName{ICCV}
\def\confYear{2025}

\title{
Controllable 3D Placement of Objects with 
Scene-Aware Diffusion Models
}

\author{
Mohamed Omran \qquad Dimitris Kalatzis \qquad Jens Petersen \qquad Amirhossein Habibian \qquad Auke Wiggers \qquad \\
Qualcomm AI Research\thanks{Qualcomm AI Research is an initiative of Qualcomm Technologies, Inc.}\\
\small{\texttt{\{momran,dkalatzi,jpeterse,auke,ahabibia\}@qti.qualcomm.com}}
}

\begin{document}


\maketitle
\begin{abstract}
Image editing approaches have become more powerful and flexible with the advent of powerful text-conditioned generative models. However, placing objects in an environment with a precise location and orientation still remains a challenge, as this typically requires carefully crafted inpainting masks or prompts.
In this work, we show that a carefully designed visual map, combined with coarse object masks, is sufficient for high quality object placement.
We design a conditioning signal that resolves ambiguities, while being flexible enough to allow for changing of shapes or object orientations.
By building on an inpainting model, we leave the background intact by design, in contrast to methods that model objects and background jointly.
We demonstrate the effectiveness of our method in the automotive setting, where we compare different conditioning signals in novel object placement tasks. These tasks are designed to measure edit quality not only in terms of appearance, but also in terms of pose and location accuracy, including cases that require non-trivial shape changes.
Lastly, we show that fine location control can be combined with appearance control to place existing objects in precise locations in a scene.
\end{abstract}
\vspace{-10pt}

\section{Introduction}

Creating 3D visual content is an essential task in many domains, with applications from robotics simulation to game development. 
In autonomous driving for example, vehicle simulators previously required users to create visual 3D assets and place these in a scene \cite{shah2018airsim, dosovitskiy2017carla},  
often a costly and manual process.
This method for synthetic data creation is now being replaced by more flexible, machine learning driven workflows \cite{yang2023unisim, hu2023gaia1}, which synthesize visual data for robustness testing or training data augmentation.
To test the robustness of a 3D detector for example, we will want to test it on objects with many different orientations.
Although image editing is now made easy by text-conditioned generative models \citep{rombach2022stablediffusion, paintbyinpaint}, the task is more challenging when the edits must adhere to such 3D instructions.
To create this data, we need fine control over object orientations and location, while retaining high image quality.



\begin{figure}[t]
    \centering
    \includegraphics[width=\linewidth]{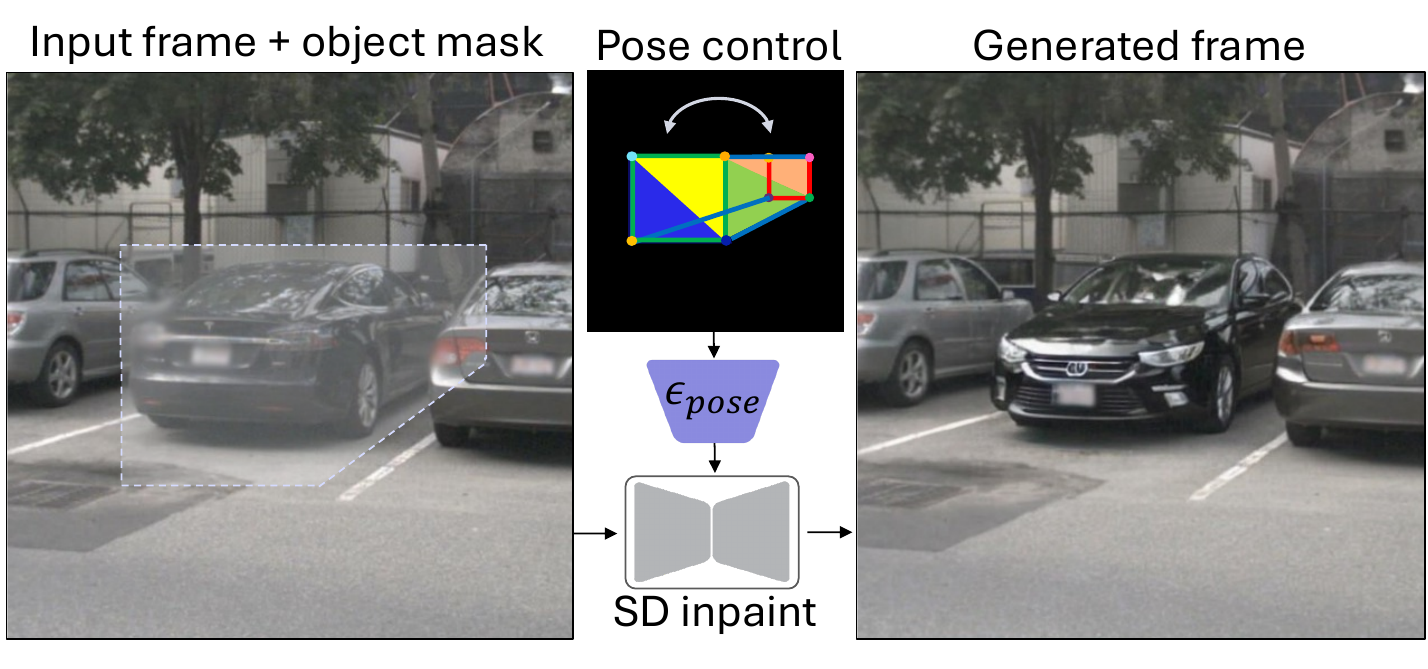}
    \caption{Overview of the use case and method. 
    We specify a target location and pose (via a visual box-map) as conditioning for an inpainting method, and protect foreground objects via their instance masks. 
    This allows us to change the orientation and shape of specific objects while keeping the rest of the scene intact.}
    \label{fig:overview}
\vspace{-10pt}
\end{figure}

Text-conditioned diffusion models, which are by now the cornerstone of many image and video editing approaches, have already achieved promising results in 3D asset creation tasks \citep{liu2023zero123, qian2023magic123, liu2024one2345, poole2022dreamfusion, lin2023magic3d}.
Still, placing these assets in existing scenes in a realistic way is not straightforward.
Various works therefore aim to control the sampling process through high-level conditioning, jointly generating the background and the objects in the desired composition.
One approach is to encode the desired location of objects in prompt-like coordinate tokens \citep{gligen, zheng2023layoutdiffusion, geodiffusion, gao2023magicdrive, wu2024neuralassets}, and generate the full frame, which ensures the result looks realistic.
However, this has only been demonstrated on low-resolution images, and it is not easy to add new objects to existing scenes as foreground and background are generated jointly.
It also requires the generative model to learn an implicit conversion from coordinate tokens to the image plane, which typically requires training on a large set of paired samples \cite{liu2023zero123}.

A different approach is to condition on desired high-level information via a visual map, such as depth \cite{controlnet, bhat2023loosecontrol, mou2024t2iadapter} or 2D layout images \cite{kim2023densediffusion}, which allows using state-of-the-art image-to-image models such as ControlNet \cite{controlnet}. 
These visual maps provide dense control, but they can be ambiguous and rigid.
For example, rotating an object in an existing depth map requires highly non-trivial modification.


In this work, we therefore propose a scalable approach for precise 3D placement of objects that uses the best of both worlds. 
We visually encode the desired location and pose by projecting a target location -- parametrized via a 3D bounding box -- onto the image plane, and use the resulting map as conditioning for generation.
We design this conditioning to be non-ambiguous with respect to object orientations, even when it is only partially visible.
This allows us to use strong image-to-image models to control generation, resulting in a highly controllable object placement method.
We show an overview of this method in Figure \ref{fig:overview}.

Through a benchmark task for 3D object placement in automotive scenes, we show that our approach results in high pose fidelity and realism, while keeping the background intact. Lastly, we combine our approach with a visual encoder to control both object appearance and location, enabling placement of specific objects in specific locations. 


\section{Related Work}

\paragraph{Diffusion-based image editing.}

Text-to-image models have been a key component in advanced image editing tools due to their flexibility and image quality \cite{rombach2022stablediffusion, hertz2022prompt}.  
We can now readily alter the visual appearance of existing objects using text instructions \cite{tumanyan2023plugandplay, hertz2022prompttoprompt, huberman2024ddpminversion}, move objects \cite{avrahami2024diffuhaul}, or insert specific objects based on one or more example images \cite{song2023objectstitch, yang2023paintbyexample, winter2024objectdrop}.
Although users can specify the desired location of objects-to-place via an inpainting mask, it is still a challenge to add objects in specific orientations, in part because text is an ambiguous medium and masks can be ambiguous as well.
Although object insertion methods that can be controlled via text prompts do exist \cite{brooks2023instructpix2pix, paintbyinpaint}, they are unreliable, and often alter the image in undesirable ways  \cite{yun2024generativelocationmodeling}.

\paragraph{Creating 3D assets.}

Text-to-image models have been extended to create 3D assets directly from text or example images.
A body of work inspired by Zero-123 \cite{liu2023zero123} lets the generative model act as a renderer \citep{liu2023zero123, liu2024one2345, tang2023makeit3d, qian2023magic123}, specifying the desired pose as an input vector.
By training a Neural Radiance Field (NeRF) on generated images from different viewpoints, we then obtain a consistent 3D asset that can be rendered from any viewpoint \cite{poole2022dreamfusion}.
Another approach is to generate full 3D assets from text or image conditioning directly \citep{chen2024text, yi2023gaussiandreamer}.
However, it is not always straightforward to use the generated assets in the image editing context, as placing them in an real background scene may require object compositing to get a realistic result.

\paragraph{Precise object placement.}

Several works aim to generate object and background jointly, which allows controlling the object locations while generating a matching background.
GLIGEN \cite{gligen} is one of the first works to show image generation conditioned on 2D bounding boxes that determine object locations.
A common strategy for 2.5D control is to render a depth map or rectangular box, and finetuning an adapter model on the resulting conditioning
\citep{bhat2023loosecontrol, eldesokey2024build}.
However, any 2D or 2.5D representation is inherently limited, as it does not capture the orientation of objects.

GeoDiffusion \citep{geodiffusion} and MagicDrive \citep{gao2023magicdrive} demonstrate image generation conditioned on 3D bounding box locations, by encoding the desired 3D box coordinates in prompt-like coordinate tokens.
This results in control over object pose, but these works only show results for low resolution images, and do not separate foreground and background.
This makes it impossible to for example add one object to an existing scene, as each inference will result in a different frame altogether.

An attempt at modeling foreground and background separately is made by Neural Assets \citep{wu2024neuralassets}, which learns to encode object appearance and location in one prompt-like vector representation, and background in another.
As this method generates the full frame, it can entangle the desired pose and background, especially when the dataset contains camera movement: moving the object may cause the background to change with it.
This method does allow placement of multiple assets in real backgrounds, but again only shows results for low ($256\times256$) resolution frames. 

%

\section{Method}
\label{sec:method}

In this work, we treat object placement as an inpainting task with coarse masks.
We specify the desired location and orientation using a visual control signal, designed to resolve ambiguity.
This results in realistic generations at high resolution, while leaving the background intact by design. 
We show a visualization of the approach in Fig.~\ref{fig:overview}.



\paragraph{Encoding object pose.}


To specify the desired location and pose for an object to generate, we create a visual conditioning map with spatial cues from the 3D bounding box.
Each 3D bounding box is a cuboid with 6 faces and 8 vertices. 
We fill each face with two triangles of distinct colors, color each vertex, and color each edge.
We then project this cuboid onto the image plane to form our conditioning.
The resulting visual map clearly indicates which side is facing forward and where on the image plane the object should be placed.
Using two triangles to color each face, as opposed to one solid color, helps determine if the object is fully or partially visible.
This input is processed by a pose encoder $\epsilon_\text{pose}$, which we implement as a ControlNet \cite{controlnet}.

\begin{figure*}[t]
     \centering
     \begin{subfigure}[b]{0.17\textwidth}
         \centering
         \includegraphics[width=\textwidth]{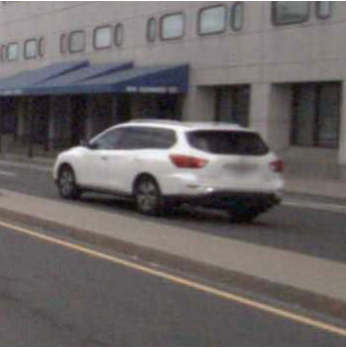}
         \caption{Original image}
         \label{fig:cond_image}
     \end{subfigure}
     \hfill
     \begin{subfigure}[b]{0.17\textwidth}
         \centering
         \includegraphics[width=\textwidth]{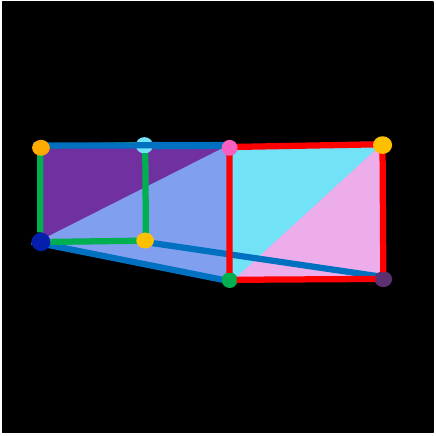}
         \caption{Bbox3d map}
         \label{fig:cond_bbox_clr}
     \end{subfigure}
     \hfill
     \begin{subfigure}[b]{0.17\textwidth}
         \centering
         \includegraphics[width=\textwidth]{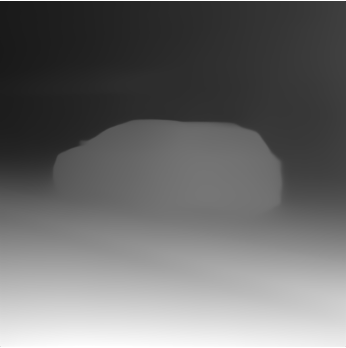}
         \caption{Depth map}
         \label{fig:cond_depth}
     \end{subfigure}
     \hfill
     \begin{subfigure}[b]{0.17\textwidth}
         \centering
         \includegraphics[width=\textwidth]{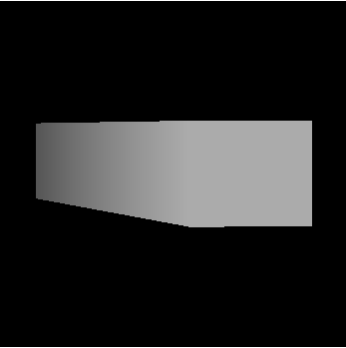}
         \caption{Bbox3d-depth map}
         \label{fig:cond_bbox_depth}
     \end{subfigure}
     \hfill
     \begin{subfigure}[b]{0.17\textwidth}
         \frame{\centering
         \includegraphics[width=\textwidth]{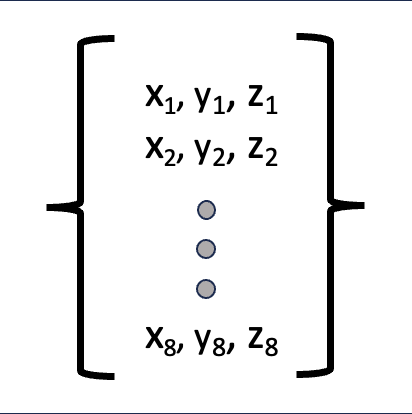}}
         \caption{Bbox coordinates}
         \label{fig:cond_coords}
     \end{subfigure}

        \caption{Various conditioning signals for 2.5D and 3D location control based on depth maps or 3D bounding boxes.}
        \label{fig:conditioning}
\vspace{-10pt}
\end{figure*}


We visualize our control signal and other methods in Fig.~\ref{fig:conditioning}. 
Depth maps (Fig.~\ref{fig:cond_depth}) are a powerful dense conditioning signal \cite{controlnet}, but with three shortcomings that our scheme addresses: (1) A depth map entangles pose and shape information, making it difficult to replace an object while changing its shape. (2) For certain objects or poses (e.g. buses viewed frontally or from the rear), a depth map is ambiguous. (3) For inserting novel objects, a depth map is more tricky to synthesize compared to a bounding box. 

We can also render a bounding box to a depth map (Fig.~\ref{fig:cond_bbox_depth}), which is the scheme similar to that of \citet{bhat2023loosecontrol}. 
This gives precise location control, but does not resolve ambiguities or provide pose control.
Finally, we could use prompt-like tokens to encode location instead (Fig.~\ref{fig:cond_coords}) \citep{geodiffusion, gao2023magicdrive}. 
Here, the downside is that the network has to learn to (implicitly) convert from token-space to image-space, which may cost training time or network capacity. 

In this work choose to make the target pose explicit by projecting the box on the image plane.
We ablate the choice of conditioning in our experiments, and demonstrate the flexibility and reliability of our approach.

\paragraph{Inpainting masks.} 
Access to detailed instance masks typically boosts the performance of inpainting models. 
However, we create coarse instance masks from the 3d bounding box here, as this allows us to place objects in new locations, or replace one object by another with a different shape. 
We obtain the mask by projecting the faces of a 3D bounding box onto the image plane, and take special care to avoid inpainting over existing objects by subtracting the instance masks of objects that occlude the box. 
The exact procedure and visual examples are given in the Appendix.



\paragraph{Inpainting model.}

After creating conditioning and masks, we can learn an editing model. 
We build on a pretrained StableDiffusion inpainting model \citep{rombach2022stablediffusion}, \texttt{stablediffusion-2-inpainting}.
Although inpainting models are sometimes less performant than models trained for generation, and inpainting is also possible by using masking in latent space, we found latent masking to result in poor realism in early experiments.
We finetune a ControlNet \citep{controlnet} to condition the inpainting model on our visual location map.
To mitigate domain shift, we finetune the UNet decoder and ControlNet parameters jointly. 
This way, the UNet can learn to generate in-domain samples, and close the domain gap between the pretraining data and our dataset of interest, while the ControlNet only needs to control pose and location of the generated object.
The inpainting mask is always the aforementioned occlusion-aware coarse mask derived from the 3D bounding box.
The resulting setup is shown in Figure \ref{fig:overview}.


Unlike previous work operating on low resolution frames \citep{geodiffusion, gao2023magicdrive, wu2024neuralassets}, working with fixed-size crops for each object allows us to edit objects at a canonical resolution.
Of course, generating a full frame and multiple objects in one inference pass can be more computationally efficient than generating objects one-by-one, but we believe that increasing inference compute proportionally to the number of objects in the frame is a desirable feature.
For more details on training and inference, and the exact setup (e.g. how we obtain per-object crops), see the Appendix. 

\section{Experiments}

\subsection{Setup}

\paragraph{Dataset.}

We report metrics on nuScenes \cite{caesar2020nuscenes}, a large-scale 3D object detection dataset, with 3D bounding box annotations for ten object categories. 
e focus on three rigid object categories that are well-represented in the dataset: cars, trucks, and buses. These objects have canonical orientations, and pose fidelity is straightforward to quantify, in contrast to objects such as traffic cones.

The nuScenes dataset contains $168.780$ training images and $36.114$ validation images. 
From the training set, we extract instances that match our categories of interest, and filter by visibility using the nuScenes toolkit with a minimum value of $3$ -- this indicates more than 60\% of the object is visible in one of the camera views at a specific timestamp.
This results in around 170.000 training instances.
We use these to train our generators and baselines.

\paragraph{Benchmark task.}

To measure accuracy of the generated objects, we use a pretrained monocular 3D detector (EPro-PnP-v2 \cite{epropnpv2}) and compare predictions on the original and edited images.
We first select data from the nuScenes validation set with visibility setting of $3$ or higher.
We then filter this set based on detector scores, to ensure that all considered instances can be detected with high precision.
This results in 27,349 instances.
We then select a subset of 5.000 instances for validation purposes, and perform editing tasks and compute all fidelity metrics on this set.
We use a second (disjoint) set of 5.000 instances as the target set for distribution-based metrics.

After editing the selected instances, we compare the detector prediction on the original frame to the prediction on the edited frame.
Following nuScenes evaluation protocol, we compute the location error as the L2-norm between predicted box centers,
and yaw error as the the top-down orientation error,
and report the mean Average Translation Error (mATE) and mean Average Orientation Error (mAOE) as the average over classes \citep{caesar2020nuscenes}.
If the yaw error is larger than 90 degrees, we say the object is ``flipped'', \textit{i.e.} it is facing the wrong direction.
To measure realism, we report the Frechet Inception Distance \cite{fid_metric}, using a second (disjoint) set of 5.000 instances as the target set.
Since we edit objects in high-resolution background frames, we need to avoid the (real) background dominating the FID computation.
We therefore take square crops around the object of interest, and compute FID on the resulting sets of crops.
The goal of this task is to keep the mAOE, mATE and ``flips'' as low as possible, while minimizing FID.
More details on this benchmark task are given in the supplementary material. 


\vspace{-4pt}
\paragraph{Baselines.}


We train five baselines with varying levels of pose-awareness, for the conditioning signals from Fig.~\ref{fig:conditioning}.
As naive baseline without pose information, we finetune the UNet of a StableDiffusion 2 inpainting model. 
Second, we train a Depth-ControlNet on depth maps extracted with an off-the-shelf monocular depth estimation method (Depth-Anything \cite{depthanything}). 
This is a strong baseline in replacement tasks, but the depth map does not allow change in shape, orientation or class, and is not easily modified.
Third, we train a ControlNet with conditioning similar to that of LooseControl \cite{bhat2023loosecontrol}, as this does allow non-trivial shape change.
To create this depth map, we render 3d bounding boxes without texture and read out the z-buffer values. 
Unlike LooseControl, this bounding box is not embedded in a scene depth map, but the conditioning signal is the same otherwise.

To compare to bounding box coordinate conditioning, we train two 3D-aware baselines.
First, we extend the coordinate conditioning of GLIGEN \citep{gligen} to accept 3D bounding box coordinates by projecting all 8 corner vertices to 2D, and using the resulting $8 \times 2$ coordinates as input.
Finally, we train a model with the MLP-based pose encoding of Neural Assets \cite{wu2024neuralassets}, which projects the 3D points to the camera plane but retains the depth coordinate, resulting in a $8 \times 3$ pose input. 
GLIGEN and Neural Assets both operate at the frame level, and require much more training compute as a result.
To enable direct comparison of pose encoding mechanisms, we train models to produce only one object.
However, as parsing pose tokens may be a more challenging task, we train the GLIGEN model for 500,000 iterations, and the Neural Assets model for 400,000 iterations.

We always finetune the pose encoder and UNet decoder jointly, for 300,000 steps (except for GLIGEN and Neural Assets), with a batch size of 4, using the AdamW optimizer and a learning rate of $1e^{-5}$.
For inference, we use a DDIM scheduler and $30$ reverse steps, with a classifier-free guidance factor $7.5$.
We use text prompts in the form \texttt{`an image of a <object class>'} during training and inference, this resulted in slightly better FID than the prompt \texttt{`<object class>`}.
Unless mentioned otherwise, we report the average score over three random seeds.
For more implementation details, see the Appendix.

\subsection{Results}

\paragraph{Object replacement.} \label{results:placement}

We first measure performance in the \emph{replacement} setting, where the goal is to replace the original object in the frame with an object that has the same size and orientation.
Results are shown in Tab.~\ref{tab:object_editing}.
The included oracle result is obtained by running the 3D detector on the original data, and computing FID between two distinct subsets of nuScenes data.

For methods that do not have orientation instructions (SD-Inpaint, LooseControl) the orientation will largely depend on the context given in the frame, and we see that these approaches obtain high orientation error (mAOE).
Our approach proves better at orientation than all baselines, and we only see large orientation error in 1\% of cases. 
CN-depth is a close second, with the best location fidelity (mATE) overall, likely because a depth map is a precise indicator for the object location.
In comparison to SD-Inpaint, the pose encoders of GLIGEN and Neural Assets reduce orientation error, but both are less precise than methods that use visual conditioning.
This may indicate that learning to parse 2D projections of the 3D bounding box corners is a more difficult task that requires longer training.

Following standard practice in object editing evaluation \cite{liu2023zero123, michel2024object}, we also measured CLIP score for all models. However, the evaluated models perform roughly the same, likely because the text prompt only describes the object category. We thus omit CLIP scores here, but show these and 3D detector confidence in the Appendix.


\begin{figure*}[t]
\centering
\includegraphics[width=0.97\linewidth]{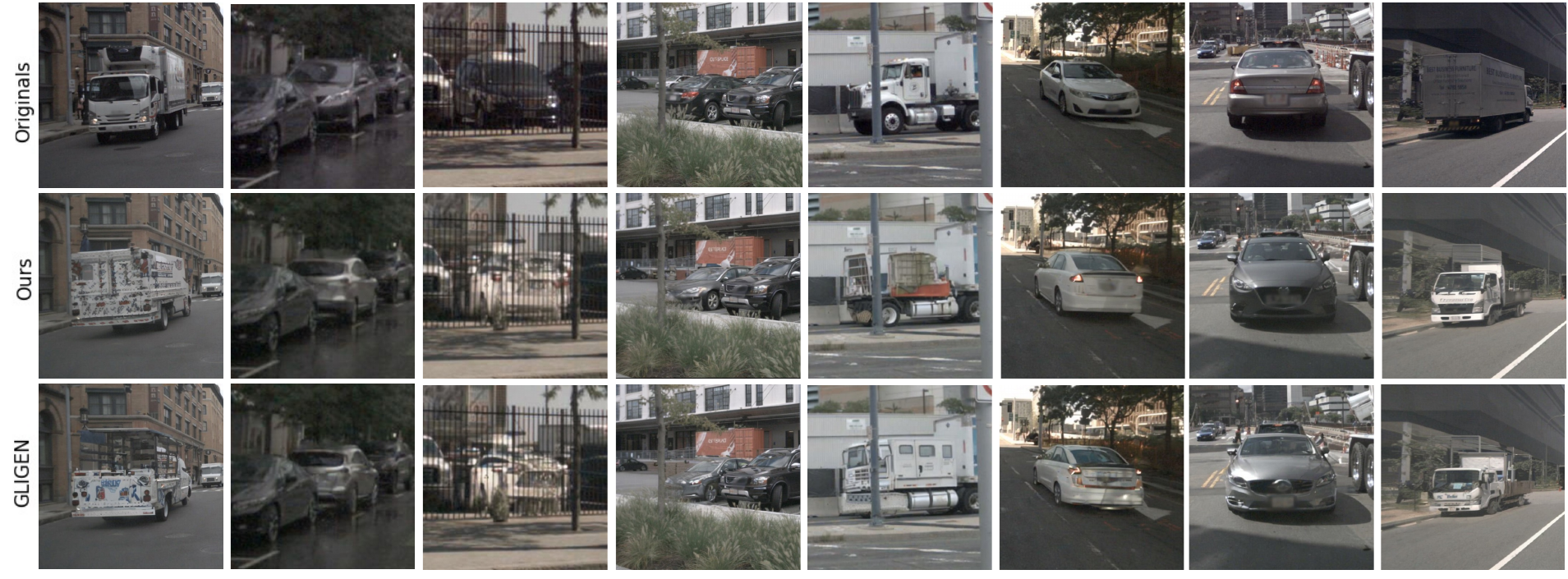}
\caption{Object replacement examples with flipped orientation instruction for various scenes, vehicles and occlusion levels. Shown here are models with explicit orientation instruction: our proposed model and GLIGEN with 3D coordinate conditioning.}
\label{fig:replacement_flipped_comparison}
\vspace{-2mm}
\end{figure*}

\vspace{-4pt}
\paragraph{Object replacement with flipping.}


Next, we measure performance in a simple modification of the replacement task, where the editing instruction is to ``flip'' the object, \textit{i.e.}, rotate it around the yaw axis by $180$ degrees. 
This often requires a non-trivial shape change and filling in previously occluded areas, as for example the back of a car may occlude the background in a different way than the front.
Results for this experiment are shown in Tab.~\ref{tab:object_editing_flipped}.
We again obtain the best mAOE, while obtaining competitive mATE and FID, showing the usefulness of the visual conditioning.
We show corresponding visual results in Fig.~\ref{fig:replacement_flipped_comparison}, including a few challenging visual occlusion cases.
GLIGEN-conditioning and our conditioning generally succeed in specifying the target orientation in these images, confirming our quantitative numbers.
For completeness, we add results for baselines that cannot change object orientation.
The depth controlnet has low mATE and FID as the depth map outlines where the object should be placed, but we cannot change object orientation, and the mAOE clearly shows the limitations of rigid conditioning.

\setlength{\tabcolsep}{4pt} 

\begin{table}[t!]
\centering

\footnotesize{

\caption{
Object editing performance in the replacement task. 
Oracle results are obtained on groundtruth data.
FID is computed with respect to a hold-out set of 5,000 objects.
The best number is shown in \textbf{bold}, second best is \underline{underlined}.
}
\vspace{-2pt}
\label{tab:object_editing}
\def\arraystretch{1.2}


\begin{tabular}{lrrrrrr}
\toprule
Method            & Steps &  mAOE$\downarrow$ & Flips$\downarrow$ & mATE$\downarrow$  & FID$\downarrow$  \\
\midrule                   
Oracle                                          &        & 0.016 & 0.00 & 0.353 & 5.60 \\
\midrule                       
SD-Inpaint \cite{rombach2022stablediffusion}    & 300k & 0.760 & 0.235 & 1.397 & 10.47 \\
LooseControl  \citep{bhat2023loosecontrol}     & 300k & 0.873 & 0.274 & 1.544 & 9.54 \\
GLIGEN enc. \cite{gligen}                      & 500k & 0.196 & 0.027 & 1.430 & 16.70 \\
Neural Assets enc. \cite{wu2024neuralassets}   & 400k & 0.198 & 0.026 & 1.543  & 11.86 \\
CN-Depth   \cite{controlnet}                   & 300k & \underline{0.133} & \underline{0.017} & \textbf{0.784} & \textbf{7.99} \\
Ours                                           & 300k & \textbf{0.121} & \textbf{0.014} & \underline{1.390} & \underline{9.36}  \\
\bottomrule
\end{tabular}



}
\vspace{-10pt}
\end{table}


\vspace{-6pt}
\paragraph{Object placement.}

Finally, we measure performance in a placement task, which involves placing vehicles in new locations.
Selecting suitable locations at scale is challenging for especially the side cameras, as they often capture an environment where cars are not usually placed, such as buildings or sidewalks. 
We therefore sample 200 uncluttered frames from the front camera of the nuScenes dataset. 
We then randomly sample three locations in the drivable space in front of the ego vehicle, which we determine using the nuScenes toolkit.
For each location, we use a fixed set of 8 orientations at 45 degree increments, and use the generative model to place a car in that location.
This results in 4,800 unique generations. We apply a 3D detector to these generated frames, and measure edit quality by comparing the output of the detector with the placement instruction.

\begin{table}[t!]
\centering

\footnotesize{

\caption{
Object editing performance with flipped orientation instruction. 
SD inpaint, CN-Depth, LooseControl have no explicit orientation instructions, but are added for completeness.
The best number is shown in \textbf{bold}, second best is \underline{underlined}.
}
\label{tab:object_editing_flipped} 
\vspace{-2pt}
\def\arraystretch{1.2}

\begin{tabular}{lrrrrr}
\toprule
Method            &  Steps & mAOE$\downarrow$ & Flips$\downarrow$ & mATE$\downarrow$  & FID$\downarrow$  \\
\midrule        

SD-Inpaint    \cite{rombach2022stablediffusion}  & 300k & 2.381   & 0.760  & 1.397 & 10.47 \\
LooseControl  \citep{bhat2023loosecontrol}       & 300k & 2.259  & 0.725 & 1.544 & \underline{9.59} \\
CN-Depth      \citep{controlnet}                 & 300k & 3.008  & 0.983 & \textbf{0.784} & \textbf{7.99} \\
GLIGEN enc.       \citep{gligen}                 & 500k & 0.494  & 0.088 & 1.607 & 17.24 \\
Neural Assets enc. \cite{wu2024neuralassets}     & 400k & \underline{0.387} & \underline{0.081} & 1.711 & 12.97 \\
Ours                                             & 300k & \textbf{0.364} & \textbf{0.073} & \underline{1.507} & 9.82 \\
\bottomrule
\end{tabular}

}
\vspace{-3pt}
\end{table}

\setlength{\tabcolsep}{4pt} 
\begin{table}[t!]
\centering

\footnotesize{
    \caption{
    Object placement performance for our method and baselines that provide precise spatial control over generations.
    The best number is shown in \textbf{bold}, second best is \underline{underlined}.
    }
    \label{tab:object_placement}
    \centering
    \def\arraystretch{1.2}

    
    
    \begin{tabular}{lrrrrr}
        \toprule
        Method & Steps & \hspace{-1mm}mAOE$\downarrow$ & Flips$\downarrow$ & \hspace{-0.5mm}mATE$\downarrow$  & FID$\downarrow$  \\
        \midrule                   
        LooseControl \cite{bhat2023loosecontrol}     & 300k & 1.575 & 0.502 & \underline{4.349} & 46.42 \\
        GLIGEN enc. \cite{gligen}                    & 500k & \underline{0.230} & \textbf{0.029} & 4.407 & 47.03 \\
        Neural Assets enc. \cite{wu2024neuralassets} & 400k & 0.253 & 0.050 & 5.340 & \textbf{45.00} \\
        Ours                       & 300k & \textbf{0.220} & \underline{0.042} & \textbf{4.512} & \underline{45.99} \\
        \bottomrule
    \end{tabular}
}
\vspace{-10pt}
\end{table}

\begin{figure*}[t]
\centering
\includegraphics[width=\linewidth]{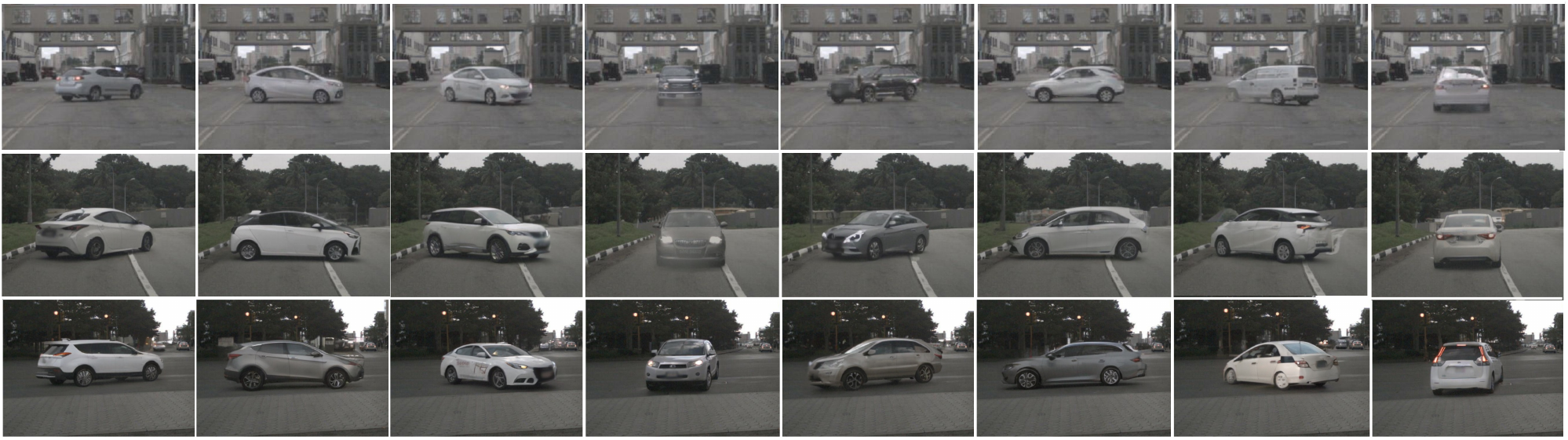}
\caption{Samples from our method for the object placement task, showing (zoomed in) editing results for 8 target orientations for three example locations. Editing is performed with text prompt and strength $1.0$, and object identity is therefore not preserved between views.}
\label{fig:placement_examples_reel}
\vspace{-8pt}
\end{figure*}

In Fig.~\ref{fig:placement_examples_reel}, we show sample results from the placement task.
We show performance in Tab.~\ref{tab:object_placement}, where we compare our method against others with location control:
the pose-aware encoders of GLIGEN and Neural Assets, and a ControlNet with LooseControl style conditioning.
In this setting, we outperform all baselines in orientation error, and GLIGEN is a close second.
LooseControl is best on translation error, but makes errors on orientation as the control signal is ambiguous.
Neural Assets performs less well, potentially indicating that the implicit conversion from coordinate tokens to the image is more challenging for this task.

\vspace{-7pt}
\paragraph{Ablation and analysis.}

We are mainly interested in object orientation, but mAOE is an aggregate metric that averages over classes.
To better understand the performance, we show per-class AOE scores for the replacement task in Tab.~\ref{tab:class_ablation_aoe}, comparing against depth-controlnet.
These results indicate that explicitly encoding the pose method is especially useful for less common vehicle types, for which orientation is sometimes hard to infer from a depth map alone.

\begin{figure}[t!]
    \centering
    \includegraphics[width=0.85\linewidth]{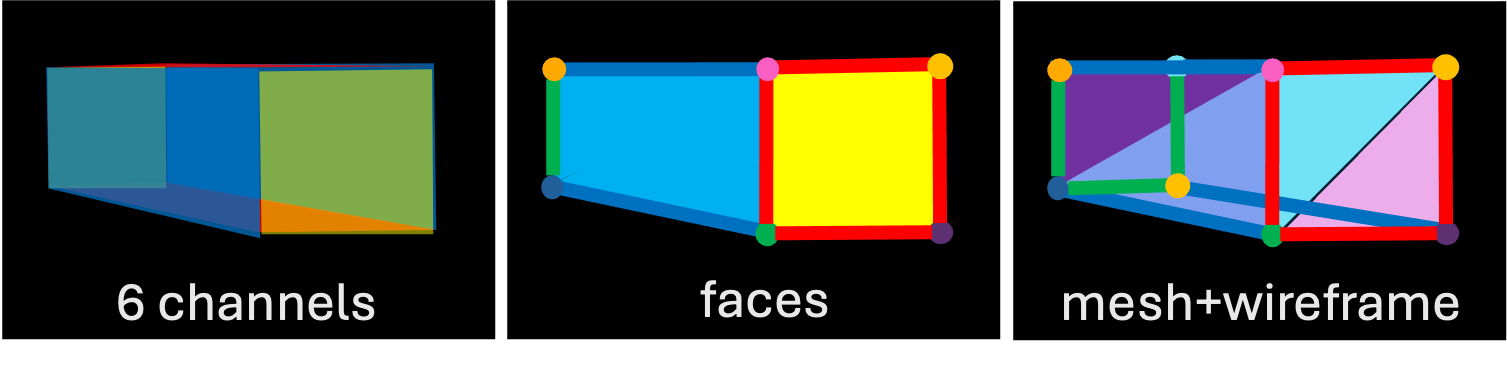}
    \vspace{-2mm}
    \caption{Three different bbox3d map conditioning signals.}
    \label{fig:visualablation}
\vspace{-10pt}
\end{figure}

Lastly, we ablate the effect of variation in the visual conditioning map by training ControlNets on the input maps shown in Fig.~\ref{fig:visualablation}.
The first variation, ``6 channels'', encodes each of the six faces in a separate channel, allowing clear delineation of where objects end, but does not resolve front/back ambiguity.
The second variation, ``faces'', only visualizes the faces visible from the ego point of view in a three channel image, which ensures the model knows which side we face, but does not specify if we see the full object or only part of it.
The third variation, ``mesh+wireframe'', uses two triangles per face and a wireframe to resolve these ambiguities, and this is the version used throughout this work.
We evaluate each network in the replacement task for three random seeds, and report the mean result in Tab.~\ref{tab:visualablation}.
All conditioning signals are fairly strong indicators for position and orientation, but we see that the wireframe conditioning results in an improvement in orientation (mAOE) and realism (FID), albeit at the cost of position accuracy (mATE).

\begin{table}[t!]
\centering

\footnotesize{
    \caption{
    Per-class Average Orientation Error (in radians) and Average Translation Error (in meters) for vehicle classes in our validation set, on the object replacement task, for one random seed.
    }
    \label{tab:class_ablation_aoe}
    \centering
    \def\arraystretch{1.2}

    \begin{tabular}{lrrrrrr} 
        \toprule 
        Category    & \multicolumn{2}{c}{cars} & \multicolumn{2}{c}{trucks} & \multicolumn{2}{c}{buses} \\ 
        \#instances & \multicolumn{2}{c}{3871} & \multicolumn{2}{c}{905} & \multicolumn{2}{c}{224}  \\ 
        \midrule
                  & AOE           & ATE           & AOE & ATE & AOE & ATE \\
        \hline                    
        GLIGEN enc. \citep{gligen}   & 0.18     & 2.96          & 0.36 & 4.41 & 0.24 & 6.04 \\ 
        Neural Assets enc. \citep{wu2024neuralassets}    & 0.18          & 3.04          & 0.31 & 4.76 & 0.27 & 6.32 \\ 
        CN-Depth  & \textbf{0.15} & \textbf{2.87} & 0.21 & \textbf{4.22}  & 0.24 &5.79 \\ 
        Ours      & \textbf{0.15} & 2.97          & \textbf{0.19} & 4.64 & \textbf{0.16} & \textbf{5.78} \\ 
        \bottomrule 
    \end{tabular}
}
\vspace{-10pt}
\end{table}


%

\subsection{Modifying object appearance.}

We show qualitatively that we can change the shape of an object, a task that can be challenging with existing editing approaches.
We also show that we can preserve object identity using a (visual) exemplar-encoder.

\vspace{-3pt}
\paragraph{Changing object category and shape.}

\begin{table}[t!]
    \centering
    \footnotesize
    \caption{Ablating the effect of bbox3d conditioning.}
    \vspace{-1mm}
    
        \begin{tabular}{lrrrrrr}
        \toprule
           Method                 & Steps &  mAOE$\downarrow$ & Flips$\downarrow$ & mATE$\downarrow$  & FID$\downarrow$  \\
        \midrule                   
            (a) six channels       & 300k & 0.124 & 0.017 & 1.414 & 10.45 \\
            (b) faces              & 300k & 0.151 & 0.024 & 1.474 &  9.86  \\
            (c) mesh+wireframe     & 300k & 0.121 & 0.014 & 1.390 &  9.36  \\
        \bottomrule
        \end{tabular}
    \label{tab:visualablation}
\end{table}

\begin{figure*}[t]
\vspace{2mm}
\centering
\includegraphics[width=1.0\linewidth]{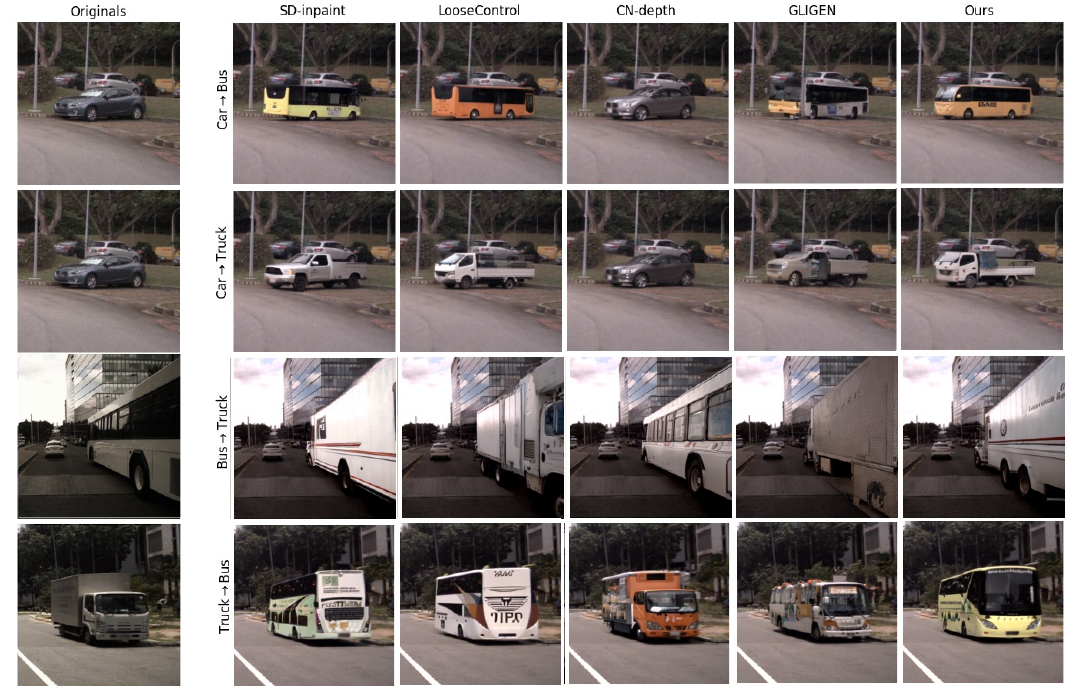}
\caption{
Example generations from our method and baseline methods, changing class but preserving orientation. 
Only GLIGEN~\cite{gligen} and our approach are able to correctly generate vehicles with the desired class change and object orientation. 
For the Depth-ControlNet (\emph{CN-depth}), most appearance information is encoded in the depth map, and it fails to modify the shape of the original object.
In contrast, our method can easily modify the class and shape of the generated object.
}
\label{fig:shapechange_examples}
\end{figure*}

In Fig.~\ref{fig:shapechange_examples}, we show the result of different editing instructions on three example inputs, where we change the class of the object while preserving its orientation.
SD-inpaint and LooseControl produce good-looking results, but come without orientation control.
CN-depth provides strong conditioning through the depth map, and the text prompt is largely ignored if it does not match the shape conditioning.
GLIGEN and our method are able to control the orientation well, but image quality is higher for our method, as confirmed by FID scores in Tab.~\ref{tab:object_editing}.
For more visual results, also see Appendix A.3.

\vspace{-3pt}
\paragraph{Preserving object identity.}

We have shown that we can place objects of a given category in specific locations, but this gives no control over object identity. 
For fine control over object appearance, we adopt an approach similar to PaintByExample \cite{yang2023paintbyexample}.
Instead of a text prompt, we train an encoder to create diffusion model conditioning from an \emph{exemplar}: an example image containing the target object. 

\begin{figure*}[t]
\vspace{1mm}
\centering
\includegraphics[width=1.0\linewidth]{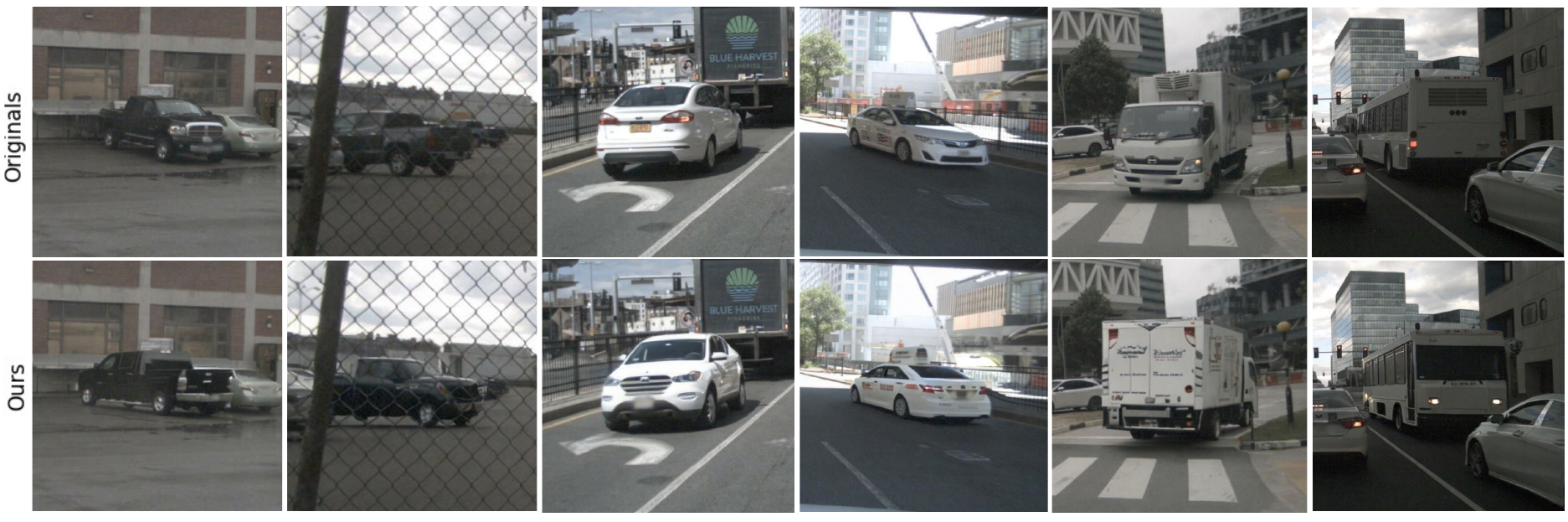}
\caption{
Samples from our method with exemplar-prompt and specific pose instruction. Top: original object frames. 
Bottom: samples with 180 degree rotation instruction for the center object. Object identity and background are preserved, while object orientation is modified.}
\vspace{1mm}
\label{fig:exemplar_rotation}
\vspace{-14pt}
\end{figure*}

Specifically, we take a DINO \cite{caron2021dinov1} encoder as frozen backbone, attach a trainable transformer block, and feed the resulting output directly to the diffusion model in place of the text prompt.
We again initialize a ControlNet, and finetune all components except the vision backbone end-to-end.
To avoid learning a trivial copying operation, we always take the exemplar from a different frame than the target frame, and augment the exemplar image by horizontal random flipping and rotation (by up to 20 degrees).

\begin{table}[t!]
\centering

\footnotesize{

\caption{
Object editing performance in replacement (top rows) and flipped replacement (bottom rows) tasks using exemplar-based prompts for one random seed.
CLIPex is the CLIP score between exemplar image and generated image, where higher is better.
}
\label{tab:exemplar}
\def\arraystretch{1.2}

\begin{tabular}{lrrrrrr}
\toprule
Method              &  \hspace{-2mm}mAOE$\downarrow$ & Flips$\downarrow$ & \hspace{-0.9mm}mATE$\downarrow$  & \hspace{-1.5mm}CLIPex$\uparrow$ & FID$\downarrow$  \\
\midrule                   
PbE \citep{yang2023paintbyexample}            & 0.456 & 0.095  & 4.502  & 72.604 & 10.82 \\
Ours                                          & 0.181 & 0.019  & 4.421  & 68.111 & 11.50 \\
\hline                                              
PbE \cite{yang2023paintbyexample} (flipped)   & 2.682 & 0.905  & 4.482  & 72.600 & 10.83 \\
Ours (flipped)                                & 0.864 & 0.195  & 4.576 & 66.641  & 11.59 \\
\bottomrule
\end{tabular}

}
\vspace{-10pt}
\end{table}

We finetune a paint-by-example checkpoint as baseline without orientation control, and show a comparison in Tab.~\ref{tab:exemplar}.
With a pose ControlNet, we can control both appearance and pose of objects, showing low mAOE for similar image quality (FID) and similarity to the exemplar (CLIPex).
Paint-by-example obtains slightly better CLIP-score between exemplar and generated image, likely because this model has been pretrained on 16 million objects and exemplars before finetuning, whereas we train our inpainting model for 300,000 steps from StableDiffusion weights. 
We show visual examples of pose control with preserved object identity in Fig.~\ref{fig:exemplar_rotation}.
For more information about this setup and additional results, see Appendix A.2.

\section{Discussion and limitations}
\label{sec:discussionandlimitations}

We present a simple conditioning scheme for object editing with precise pose and location control. 
Our scheme decouples shape from pose, allowing us to perform non-trivial object editing, \textit{e.g.}, changing object identity or flipping an object around its vertical axis, while maintaining high pose fidelity. 
Our approach also helps with objects whose shape is less indicative of pose, \textit{e.g.}, for a bus, the outline or depth map may not be sufficient to tell whether the camera faces its front or back side. 
The conditioning signal is easy to synthesize, and placing new objects in empty places only requires projection of 3D bounding boxes to the image plane.
In \cref{tab:overall_ranking}, we rank methods on pose fidelity, a property that we believe is especially important for downstream tasks, and our method consistently outperforms baselines in three different settings.

In this work, we deliberately focus on generating a single object at a time.
This makes it possible to scale to high resolution, but it does make our approach computationally less efficient than approaches that render full scenes.
We argue that this is not a limitation, as we can still render entire scenes by iterating over objects one by one.
An advantage of our approach is that we can use the same inference budget for each \emph{object}, whereas approaches such as GeoDiffusion \cite{geodiffusion} generate full frames with a fixed inference budget for the entire scene, no matter how complex that scene is.
It is reasonable to believe that a crowded scene with complex object interactions is more difficult to render, and that it may require more computational power to do so than for an empty street scene.
By rendering objects one-by-one, we adjust the compute budget for scene complexity.

A potential limitation of our approach is the reliance on high quality 3D bounding box annotations for generator training, as well as high quality camera calibration parameters.
These are readily available for common driving datasets, but this is not true for all domains. 
If ground truth bounding boxes are not available, we believe good pseudo-groundtruth can be obtained using pretrained detectors, but we have also seen that 3D detectors are not perfectly reliable, obtaining reasonably high translation error on oracle data \citep{epropnpv2}.
Nevertheless, we believe that as scene understanding algorithms advance, so too will the quality of the 3D annotations, even without ground truth data.


\section{Conclusion}

\setlength{\tabcolsep}{3pt} 
\begin{table}[t!]
\footnotesize{
\caption{Our method consistently outranks competing baselines across different tasks when it comes to pose fidelity. Since the conditioning map is easy to synthesize and manipulate it can be used for flexible edits that involve new objects or shape change.}
\def\arraystretch{1.2}
    \centering
    \label{tab:overall_ranking}
    \begin{tabular}{lccc}
        \toprule
                         & \multicolumn{3}{c}{Pose Fidelity Ranking (mAOE)} \\
        \midrule                   
        Method           & Replacement & \makecell{Replacement \\ (+ Shape Change)} & Placement \\
        \midrule                   
        SD-Inpaint \cite{rombach2022stablediffusion} & 5\ts{th} & 5\ts{th} & 4\ts{th} \\
        LooseControl \citep{bhat2023loosecontrol}    & 6\ts{th} & 4\ts{th} &   -      \\
        GLIGEN enc. \citep{gligen}                   & 3\ts{rd} & 3\ts{rd} & 2\ts{nd} \\
        Neural Assets enc. \cite{wu2024neuralassets} & 4\ts{th} & 2\ts{nd} & 3\ts{rd} \\
        CN-Depth \citep{controlnet}                  & 2\ts{nd} & 6\ts{th} &  -       \\
        Ours                                         & \textbf{1\ts{st}} & \textbf{1\ts{st}} & \textbf{1\ts{st}} \\
        \bottomrule
    \end{tabular}
}
\vspace{-8pt}
\end{table}

In this work, we propose an effective approach for 3D object editing at high resolution.
We carefully design a visual map that specifies the the desired location and orientation, while resolving ambiguities present in other visual conditioning maps.
This choice allows us to leverage state-of-the-art image-to-image models for generation.
We show that using our approach, we can place arbitrary objects in existing scenes while keeping the background scene intact.
Furthermore, we show that we can reliably perform non-trivial edits: changing the category and shape of an object while preserving its orientation, or preserving object identity while changing its orientation. 

We believe that precise 3D control of generative vision models is an important step towards neural simulators that can interact with scenes in a spatially consistent manner.
To this end, this work provides a direct comparison of control mechanisms in a controlled setting, highlighting strengths and weaknesses of the different methods.

{
    \small
    \bibliographystyle{iccv_author_kit/ieeenat_fullname}
    \bibliography{main}

\begin{thebibliography}{41}
\providecommand{\natexlab}[1]{#1}
\providecommand{\url}[1]{\texttt{#1}}
\expandafter\ifx\csname urlstyle\endcsname\relax
  \providecommand{\doi}[1]{doi: #1}\else
  \providecommand{\doi}{doi: \begingroup \urlstyle{rm}\Url}\fi

\bibitem[Avrahami et~al.(2024)Avrahami, Gal, Chechik, Fried, Lischinski, Vahdat, and Nie]{avrahami2024diffuhaul}
Omri Avrahami, Rinon Gal, Gal Chechik, Ohad Fried, Dani Lischinski, Arash Vahdat, and Weili Nie.
\newblock {D}iff{U}{H}aul: A training-free method for object dragging in images.
\newblock In \emph{SIGGRAPH Asia 2024 Conference Papers}, pages 1--12, 2024.

\bibitem[Bhat et~al.(2024)Bhat, Mitra, and Wonka]{bhat2023loosecontrol}
Shariq~Farooq Bhat, Niloy Mitra, and Peter Wonka.
\newblock Loose{C}ontrol: Lifting controlnet for generalized depth conditioning.
\newblock In \emph{ACM SIGGRAPH 2024 Conference Papers}, pages 1--11, 2024.

\bibitem[Brooks et~al.(2023)Brooks, Holynski, and Efros]{brooks2023instructpix2pix}
Tim Brooks, Aleksander Holynski, and Alexei~A Efros.
\newblock Instructpix2pix: Learning to follow image editing instructions.
\newblock In \emph{Proceedings of the IEEE/CVF Conference on Computer Vision and Pattern Recognition}, pages 18392--18402, 2023.

\bibitem[Caesar et~al.(2020)Caesar, Bankiti, Lang, Vora, Liong, Xu, Krishnan, Pan, Baldan, and Beijbom]{caesar2020nuscenes}
Holger Caesar, Varun Bankiti, Alex~H Lang, Sourabh Vora, Venice~Erin Liong, Qiang Xu, Anush Krishnan, Yu Pan, Giancarlo Baldan, and Oscar Beijbom.
\newblock nuscenes: A multimodal dataset for autonomous driving.
\newblock In \emph{Proceedings of the IEEE conference on Computer Vision and Pattern Recognition}, 2020.

\bibitem[Caron et~al.(2021)Caron, Touvron, Misra, J{\'e}gou, Mairal, Bojanowski, and Joulin]{caron2021dinov1}
Mathilde Caron, Hugo Touvron, Ishan Misra, Herv{\'e} J{\'e}gou, Julien Mairal, Piotr Bojanowski, and Armand Joulin.
\newblock Emerging properties in self-supervised vision transformers.
\newblock In \emph{IEEE International Conference on Computer Vision}, pages 9650--9660, 2021.

\bibitem[Chen et~al.(2022)Chen, Wang, Wang, Tian, Xiong, and Li]{epropnpv2}
Hansheng Chen, Pichao Wang, Fan Wang, Wei Tian, Lu Xiong, and Hao Li.
\newblock Epro-{PnP}: {G}eneralized end-to-end probabilistic perspective-n-points for monocular object pose estimation.
\newblock In \emph{Proceedings of the IEEE/CVF conference on computer vision and pattern recognition}, pages 2781--2790, 2022.

\bibitem[Chen et~al.(2024{\natexlab{a}})Chen, Xie, Chen, Wang, Hong, Li, and Yeung]{geodiffusion}
Kai Chen, Enze Xie, Zhe Chen, Yibo Wang, Lanqing Hong, Zhenguo Li, and Dit-Yan Yeung.
\newblock Geodiffusion: Text-prompted geometric control for object detection data generation.
\newblock \emph{International Conference on Learning Representations}, 2024{\natexlab{a}}.

\bibitem[Chen et~al.(2024{\natexlab{b}})Chen, Wang, Wang, and Liu]{chen2024text}
Zilong Chen, Feng Wang, Yikai Wang, and Huaping Liu.
\newblock Text-to-3d using gaussian splatting.
\newblock In \emph{Proceedings of the IEEE/CVF Conference on Computer Vision and Pattern Recognition}, pages 21401--21412, 2024{\natexlab{b}}.

\bibitem[Dosovitskiy et~al.(2017)Dosovitskiy, Ros, Codevilla, Lopez, and Koltun]{dosovitskiy2017carla}
Alexey Dosovitskiy, German Ros, Felipe Codevilla, Antonio Lopez, and Vladlen Koltun.
\newblock Carla: An open urban driving simulator.
\newblock In \emph{Conference on robot learning}, pages 1--16. PMLR, 2017.

\bibitem[Eldesokey and Wonka(2024)]{eldesokey2024build}
Abdelrahman Eldesokey and Peter Wonka.
\newblock Build-a-scene: Interactive 3d layout control for diffusion-based image generation.
\newblock \emph{arXiv preprint arXiv:2408.14819}, 2024.

\bibitem[Gao et~al.(2024)Gao, Chen, Xie, Hong, Li, Yeung, and Xu]{gao2023magicdrive}
Ruiyuan Gao, Kai Chen, Enze Xie, Lanqing Hong, Zhenguo Li, Dit-Yan Yeung, and Qiang Xu.
\newblock Magicdrive: Street view generation with diverse 3d geometry control.
\newblock \emph{International Conference on Learning Representations}, 2024.

\bibitem[Hertz et~al.(2022{\natexlab{a}})Hertz, Mokady, Tenenbaum, Aberman, Pritch, and Cohen-Or]{hertz2022prompt}
Amir Hertz, Ron Mokady, Jay Tenenbaum, Kfir Aberman, Yael Pritch, and Daniel Cohen-Or.
\newblock Prompt-to-prompt image editing with cross attention control.
\newblock \emph{arXiv preprint arXiv:2208.01626}, 2022{\natexlab{a}}.

\bibitem[Hertz et~al.(2022{\natexlab{b}})Hertz, Mokady, Tenenbaum, Aberman, Pritch, and Cohen-Or]{hertz2022prompttoprompt}
Amir Hertz, Ron Mokady, Jay Tenenbaum, Kfir Aberman, Yael Pritch, and Daniel Cohen-Or.
\newblock Prompt-to-prompt image editing with cross attention control.
\newblock \emph{arXiv preprint arXiv:2208.01626}, 2022{\natexlab{b}}.

\bibitem[Heusel et~al.(2017)Heusel, Ramsauer, Unterthiner, Nessler, and Hochreiter]{fid_metric}
Martin Heusel, Hubert Ramsauer, Thomas Unterthiner, Bernhard Nessler, and Sepp Hochreiter.
\newblock Gans trained by a two time-scale update rule converge to a local nash equilibrium.
\newblock In \emph{Neural Information Processing Systems}, 2017.

\bibitem[Hu et~al.(2023)Hu, Russell, Yeo, Murez, Fedoseev, Kendall, Shotton, and Corrado]{hu2023gaia1}
Anthony Hu, Lloyd Russell, Hudson Yeo, Zak Murez, George Fedoseev, Alex Kendall, Jamie Shotton, and Gianluca Corrado.
\newblock Gaia-1: A generative world model for autonomous driving.
\newblock \emph{arXiv preprint arXiv:2309.17080}, 2023.

\bibitem[Huberman-Spiegelglas et~al.(2024)Huberman-Spiegelglas, Kulikov, and Michaeli]{huberman2024ddpminversion}
Inbar Huberman-Spiegelglas, Vladimir Kulikov, and Tomer Michaeli.
\newblock An edit friendly ddpm noise space: Inversion and manipulations.
\newblock In \emph{Proceedings of the IEEE/CVF Conference on Computer Vision and Pattern Recognition}, pages 12469--12478, 2024.

\bibitem[Kim et~al.(2023)Kim, Lee, Kim, Ha, and Zhu]{kim2023densediffusion}
Yunji Kim, Jiyoung Lee, Jin-Hwa Kim, Jung-Woo Ha, and Jun-Yan Zhu.
\newblock Dense text-to-image generation with attention modulation.
\newblock In \emph{Proceedings of the IEEE/CVF International Conference on Computer Vision}, pages 7701--7711, 2023.

\bibitem[Kirillov et~al.(2023)Kirillov, Mintun, Ravi, Mao, Rolland, Gustafson, Xiao, Whitehead, Berg, Lo, Dollar, and Girshick]{Kirillov_2023_ICCV}
Alexander Kirillov, Eric Mintun, Nikhila Ravi, Hanzi Mao, Chloe Rolland, Laura Gustafson, Tete Xiao, Spencer Whitehead, Alexander~C. Berg, Wan-Yen Lo, Piotr Dollar, and Ross Girshick.
\newblock Segment anything.
\newblock In \emph{Proceedings of the IEEE/CVF {I}nternational {C}onference on {C}omputer {V}ision (ICCV)}, pages 4015--4026, 2023.

\bibitem[Li et~al.(2023)Li, Liu, Wu, Mu, Yang, Gao, Li, and Lee]{gligen}
Yuheng Li, Haotian Liu, Qingyang Wu, Fangzhou Mu, Jianwei Yang, Jianfeng Gao, Chunyuan Li, and Yong~Jae Lee.
\newblock {GLIGEN:} open-set grounded text-to-image generation.
\newblock In \emph{{IEEE/CVF} Conference on Computer Vision and Pattern Recognition, {CVPR} 2023, Vancouver, BC, Canada, June 17-24, 2023}, pages 22511--22521. {IEEE}, 2023.

\bibitem[Lin et~al.(2023)Lin, Gao, Tang, Takikawa, Zeng, Huang, Kreis, Fidler, Liu, and Lin]{lin2023magic3d}
Chen-Hsuan Lin, Jun Gao, Luming Tang, Towaki Takikawa, Xiaohui Zeng, Xun Huang, Karsten Kreis, Sanja Fidler, Ming-Yu Liu, and Tsung-Yi Lin.
\newblock Magic3d: High-resolution text-to-3d content creation.
\newblock In \emph{Proceedings of the IEEE/CVF Conference on Computer Vision and Pattern Recognition}, pages 300--309, 2023.

\bibitem[Liu et~al.(2024)Liu, Xu, Jin, Chen, Varma~T, Xu, and Su]{liu2024one2345}
Minghua Liu, Chao Xu, Haian Jin, Linghao Chen, Mukund Varma~T, Zexiang Xu, and Hao Su.
\newblock One-2-3-45: Any single image to 3d mesh in 45 seconds without per-shape optimization.
\newblock \emph{Advances in Neural Information Processing Systems}, 36, 2024.

\bibitem[Liu et~al.(2023)Liu, Wu, Van~Hoorick, Tokmakov, Zakharov, and Vondrick]{liu2023zero123}
Ruoshi Liu, Rundi Wu, Basile Van~Hoorick, Pavel Tokmakov, Sergey Zakharov, and Carl Vondrick.
\newblock Zero-1-to-3: Zero-shot one image to 3d object.
\newblock In \emph{Proceedings of the IEEE/CVF international conference on computer vision}, pages 9298--9309, 2023.

\bibitem[Michel et~al.(2024)Michel, Bhattad, VanderBilt, Krishna, Kembhavi, and Gupta]{michel2024object}
Oscar Michel, Anand Bhattad, Eli VanderBilt, Ranjay Krishna, Aniruddha Kembhavi, and Tanmay Gupta.
\newblock Object 3dit: Language-guided 3d-aware image editing.
\newblock \emph{Advances in Neural Information Processing Systems}, 36, 2024.

\bibitem[Mou et~al.(2024)Mou, Wang, Xie, Wu, Zhang, Qi, and Shan]{mou2024t2iadapter}
Chong Mou, Xintao Wang, Liangbin Xie, Yanze Wu, Jian Zhang, Zhongang Qi, and Ying Shan.
\newblock T2i-adapter: Learning adapters to dig out more controllable ability for text-to-image diffusion models.
\newblock In \emph{Proceedings of the AAAI Conference on Artificial Intelligence}, pages 4296--4304, 2024.

\bibitem[Poole et~al.(2022)Poole, Jain, Barron, and Mildenhall]{poole2022dreamfusion}
Ben Poole, Ajay Jain, Jonathan~T Barron, and Ben Mildenhall.
\newblock Dreamfusion: Text-to-3d using 2d diffusion.
\newblock \emph{arXiv preprint arXiv:2209.14988}, 2022.

\bibitem[Qian et~al.(2023)Qian, Mai, Hamdi, Ren, Siarohin, Li, Lee, Skorokhodov, Wonka, Tulyakov, et~al.]{qian2023magic123}
Guocheng Qian, Jinjie Mai, Abdullah Hamdi, Jian Ren, Aliaksandr Siarohin, Bing Li, Hsin-Ying Lee, Ivan Skorokhodov, Peter Wonka, Sergey Tulyakov, et~al.
\newblock Magic123: One image to high-quality 3d object generation using both 2d and 3d diffusion priors.
\newblock \emph{arXiv preprint arXiv:2306.17843}, 2023.

\bibitem[Rombach et~al.(2022)Rombach, Blattmann, Lorenz, Esser, and Ommer]{rombach2022stablediffusion}
Robin Rombach, Andreas Blattmann, Dominik Lorenz, Patrick Esser, and Bj{\"o}rn Ommer.
\newblock High-resolution image synthesis with latent diffusion models.
\newblock In \emph{Proceedings of the IEEE conference on Computer Vision and Pattern Recognition}, 2022.

\bibitem[Shah et~al.(2018)Shah, Dey, Lovett, and Kapoor]{shah2018airsim}
Shital Shah, Debadeepta Dey, Chris Lovett, and Ashish Kapoor.
\newblock Airsim: High-fidelity visual and physical simulation for autonomous vehicles.
\newblock In \emph{Field and Service Robotics: Results of the 11th International Conference}, pages 621--635. Springer, 2018.

\bibitem[Song et~al.(2023)Song, Zhang, Lin, Cohen, Price, Zhang, Kim, and Aliaga]{song2023objectstitch}
Yizhi Song, Zhifei Zhang, Zhe Lin, Scott Cohen, Brian Price, Jianming Zhang, Soo~Ye Kim, and Daniel Aliaga.
\newblock Objectstitch: Object compositing with diffusion model.
\newblock In \emph{Proceedings of the IEEE/CVF Conference on Computer Vision and Pattern Recognition}, pages 18310--18319, 2023.

\bibitem[Tang et~al.(2023)Tang, Wang, Zhang, Zhang, Yi, Ma, and Chen]{tang2023makeit3d}
Junshu Tang, Tengfei Wang, Bo Zhang, Ting Zhang, Ran Yi, Lizhuang Ma, and Dong Chen.
\newblock Make-it-3d: High-fidelity 3d creation from a single image with diffusion prior.
\newblock In \emph{Proceedings of the IEEE/CVF international conference on computer vision}, pages 22819--22829, 2023.

\bibitem[Tumanyan et~al.(2023)Tumanyan, Geyer, Bagon, and Dekel]{tumanyan2023plugandplay}
Narek Tumanyan, Michal Geyer, Shai Bagon, and Tali Dekel.
\newblock Plug-and-play diffusion features for text-driven image-to-image translation.
\newblock In \emph{Proceedings of the IEEE/CVF Conference on Computer Vision and Pattern Recognition}, pages 1921--1930, 2023.

\bibitem[Wasserman et~al.(2024)Wasserman, Rotstein, Ganz, and Kimmel]{paintbyinpaint}
Navve Wasserman, Noam Rotstein, Roy Ganz, and Ron Kimmel.
\newblock Paint by inpaint: Learning to add image objects by removing them first.
\newblock \emph{arXiv preprint arXiv:2404.18212}, 2024.

\bibitem[Winter et~al.(2024)Winter, Cohen, Fruchter, Pritch, Rav-Acha, and Hoshen]{winter2024objectdrop}
Daniel Winter, Matan Cohen, Shlomi Fruchter, Yael Pritch, Alex Rav-Acha, and Yedid Hoshen.
\newblock Objectdrop: Bootstrapping counterfactuals for photorealistic object removal and insertion.
\newblock \emph{arXiv preprint arXiv:2403.18818}, 2024.

\bibitem[Wu et~al.(2024)Wu, Rubanova, Kabra, Hudson, Gilitschenski, Aytar, van Steenkiste, Allen, and Kipf]{wu2024neuralassets}
Ziyi Wu, Yulia Rubanova, Rishabh Kabra, Drew~A Hudson, Igor Gilitschenski, Yusuf Aytar, Sjoerd van Steenkiste, Kelsey~R Allen, and Thomas Kipf.
\newblock Neural assets: 3d-aware multi-object scene synthesis with image diffusion models.
\newblock \emph{arXiv preprint arXiv:2406.09292}, 2024.

\bibitem[Yang et~al.(2023{\natexlab{a}})Yang, Gu, Zhang, Zhang, Chen, Sun, Chen, and Wen]{yang2023paintbyexample}
Binxin Yang, Shuyang Gu, Bo Zhang, Ting Zhang, Xuejin Chen, Xiaoyan Sun, Dong Chen, and Fang Wen.
\newblock Paint by example: Exemplar-based image editing with diffusion models.
\newblock In \emph{Proceedings of the IEEE/CVF Conference on Computer Vision and Pattern Recognition}, pages 18381--18391, 2023{\natexlab{a}}.

\bibitem[Yang et~al.(2024)Yang, Kang, Huang, Xu, Feng, and Zhao]{depthanything}
Lihe Yang, Bingyi Kang, Zilong Huang, Xiaogang Xu, Jiashi Feng, and Hengshuang Zhao.
\newblock Depth anything: Unleashing the power of large-scale unlabeled data.
\newblock In \emph{{IEEE/CVF} Conference on Computer Vision and Pattern Recognition, {CVPR} 2024, Seattle, WA, USA, June 16-22, 2024}, pages 10371--10381. {IEEE}, 2024.

\bibitem[Yang et~al.(2023{\natexlab{b}})Yang, Chen, Wang, Manivasagam, Ma, Yang, and Urtasun]{yang2023unisim}
Ze Yang, Yun Chen, Jingkang Wang, Sivabalan Manivasagam, Wei-Chiu Ma, Anqi~Joyce Yang, and Raquel Urtasun.
\newblock Unisim: A neural closed-loop sensor simulator.
\newblock In \emph{Proceedings of the IEEE/CVF Conference on Computer Vision and Pattern Recognition}, pages 1389--1399, 2023{\natexlab{b}}.

\bibitem[Yi et~al.(2023)Yi, Fang, Wu, Xie, Zhang, Liu, Tian, and Wang]{yi2023gaussiandreamer}
Taoran Yi, Jiemin Fang, Guanjun Wu, Lingxi Xie, Xiaopeng Zhang, Wenyu Liu, Qi Tian, and Xinggang Wang.
\newblock Gaussiandreamer: Fast generation from text to 3d gaussian splatting with point cloud priors.
\newblock \emph{arXiv preprint arXiv:2310.08529}, 2023.

\bibitem[Yun et~al.(2024)Yun, Abati, Omran, Choo, Habibian, and Wiggers]{yun2024generativelocationmodeling}
Jooyeol Yun, Davide Abati, Mohamed Omran, Jaegul Choo, Amirhossein Habibian, and Auke Wiggers.
\newblock Generative location modeling for spatially aware object insertion.
\newblock \emph{arXiv preprint arXiv:2410.13564}, 2024.

\bibitem[Zhang et~al.(2023)Zhang, Rao, and Agrawala]{controlnet}
Lvmin Zhang, Anyi Rao, and Maneesh Agrawala.
\newblock Adding conditional control to text-to-image diffusion models, 2023.

\bibitem[Zheng et~al.(2023)Zheng, Zhou, Li, Qi, Shan, and Li]{zheng2023layoutdiffusion}
Guangcong Zheng, Xianpan Zhou, Xuewei Li, Zhongang Qi, Ying Shan, and Xi Li.
\newblock Layoutdiffusion: Controllable diffusion model for layout-to-image generation.
\newblock In \emph{Proceedings of the IEEE/CVF Conference on Computer Vision and Pattern Recognition}, pages 22490--22499, 2023.

\end{thebibliography}
}

\newpage
\appendix
\setcounter{page}{1}
\maketitlesupplementary
\section{Appendix}

\subsection{Dataset}
\label{appendix:sec:dataset}

\subsubsection{Selecting instances for evaluation}
\label{appendix:sec:dataset:selecting_instances}

\paragraph{Replacement benchmark.}

We select 5000 instances from the nuScenes dataset for evaluation purposes. 
We want these instances to be clearly visible, easy to detect with off-the-shelf detectors, so we can reliably compute standard metrics for orientation and translation error, and do not have to consider the false negative case, where a detector does not make a prediction for a given object.

To arrive at these 5000 detectable instances, we apply a series of filters to objects annotated in the dataset. 
We first retain only cars, trucks, and buses, as these are rigid objects that (a) are well-represented in the dataset (unlike \textit{e.g.} trailers and construction vehicles), and (b) have a canonical back-to-front axis such that we can non-trivially determine the pose (unlike \textit{e.g.} traffic cones and barriers). 
We then remove instances that occupy an area of the image smaller than $96\times96$ pixels, or are closer than $4$ meters or further away than $40$ meters from the ego vehicle.
Small objects tend to be blurry and poorly visible, while close objects are sometimes behind the ego vehicle. 

Next, we retain only objects that are moderately occluded or unoccluded. 
Here, we rely on the occlusion labels provided by nuScenes: Objects are assigned numerical labels from 1 to 4 depending on their visibility. 
We keep objects that have a label of 3 or 4, meaning objects that are either 60-80\% visible or 80-100\% visible respectively. 
These labels are imperfect, since the visibility of an object at a specific point in time is determined based on all six camera views captured taken together at that same moment in time. 
Thus, if one half of an object is visible in the front camera and the other half is visible in one of the side cameras, it is assigned a visibility label of 4 (80-100\%).
However, this works well enough in practice, and allows us to largely avoid objects that are heavily occluded or even completely invisible in certain frames.

Finally, we filter out objects that our reference detector (EPro-Pnp-v2 \cite{epropnpv2}) fails to detect with high pose accuracy (more than $3\deg$ yaw error). 
This is to make sure that our pose fidelity results are dependent on edit quality, rather than detector error. 
Of the surviving instances, we keep a random selection of 5000 for measuring edit quality, and a disjoint set of 5000 as a reference set for measuring FID.

\paragraph{Placement benchmark.}

For evaluating object placement, we require new locations that do not contain any vehicles.
We opted for a simple solution: We randomly selected 200 images from the frontal camera, where the area in front of the ego vehicle is clear of any annotated objects. Within this area, we sample three points at varying distances. At each point, we place eight bounding boxes: First, we place a box that is aligned with the ego vehicle. Then we create seven others, obtained by successive rotations on the ground plane with increments of $45\deg$. Thus for each of the 200 images, we require 24 separate edits, resulting in a total of 4800 edits to measure placement performance.

\subsubsection{Instance masks.}
\label{appendix:sec:dataset:instancemasks}


\begin{figure*}[t]
     \hspace{10mm}
     \centering
     \begin{subfigure}[b]{0.18\textwidth}
         \centering
         \includegraphics[width=\textwidth]{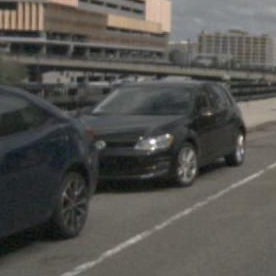}
         \caption{Original image.}
         \label{fig:image}
     \end{subfigure}
     \hfill
     \begin{subfigure}[b]{0.18\textwidth}
         \centering
         \includegraphics[width=\textwidth]{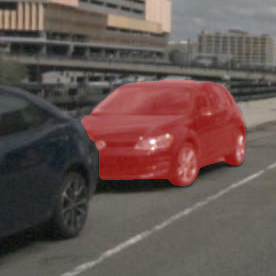}
         \caption{Fine instance mask}
         \label{fig:mask_detailed}
     \end{subfigure}
     \hfill
     \begin{subfigure}[b]{0.18\textwidth}
         \centering
         \includegraphics[width=\textwidth]{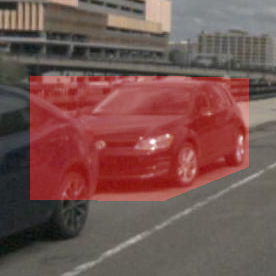}
         \caption{Coarse bbox3d mask}
         \label{fig:mask_bbox3d}
     \end{subfigure}
     \hfill
     \begin{subfigure}[b]{0.18\textwidth}
         \centering
         \includegraphics[width=\textwidth]{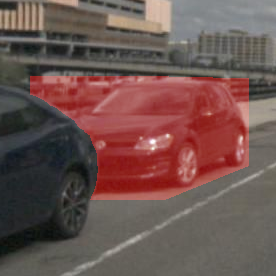}
         \caption{Occlusion-aware mask}
         \label{fig:mask_occl_aware}
     \end{subfigure}
     \hspace{10mm}
        \caption{
        Detailed inpainting masks (\ref{fig:mask_detailed}) that fit the original object exactly do not allow for replacement with non-trivial shape change. Coarse masks derived from 3D bounding boxes on the other hand (\ref{fig:mask_bbox3d}) require inpainting models to fill in occluding foreground objects besides replacing the main target object, and possibly destroy foreground details. We instead extract instance masks for all annotated objects, and use these to create an occlusion-aware 3D bounding box shaped mask (\ref{fig:mask_occl_aware}).}
        \label{fig:inpainting-masks}
\end{figure*}


To replace or insert an object, our pipeline requires a binary inpainting mask indicating which region of the image needs modifying. In the case of object replacement, we need to derive a mask for the existing (target) object. nuScenes does not provide pixel precise instance masks for annotated objects. In any case, such masks would not be suitable as inpainting masks, since they make it very difficult to replace a target object while modifying its shape (e.g. replacing a hatchback car with a sedan). Instead, we derive inpainting masks from the 3D bounding box annotations: We project the bounding box corners to 2D points and fill in the convex hull. This results in a flexible mask that allows for replacement with shape change (see \cref{tab:object_editing_flipped}, main paper). It is also possible to manipulate 3D bounding boxes prior to the mask projection step, e.g. enlarging the box to replace the target object with a bigger one.

Inpainting masks derived from 3D bounding box annotations have one downside: Our pipeline will often have to reproduce parts of the background, or -- if the target object is occluded -- also have to reproduce parts of the foreground (see \cref{fig:inpainting-masks}, main paper). To mitigate this, we resort to subtracting the pixel precise masks of occluding objects from the 3D bounding box masks of target objects. As a first step, we use an off-the-shelf instance segmentation method (SAM-v1 \cite{Kirillov_2023_ICCV}) to extract pixel-precise masks for every annotated object in nuScenes. Then, for each target object, we use a renderer to determine which objects are occluders, and subtract the masks accordingly.

This remains a partial solution, since target objects can be occluded by parts of the scene that are not annotated as stand-alone objects (e.g. buildings and lampposts). However, we observed that this step already improves pose fidelity considerably for occluded target objects.

For a visualization of different mask types, including what we refer to as an \emph{occlusion-aware mask}, see Fig.~\ref{fig:inpainting-masks}.

\subsection{Method details}
\label{appendix:sec:method}

\paragraph{GLIGEN}
GLIGEN~\cite{gligen} works by encoding bounding box coordinates with a Fourier embedding.
This embedding is concatenated with a prompt embedding that describes the object within that box (using the same text encoder as the original model).
GLIGEN can handle multiple objects and produces a token for each object in this way, but in our setup we only ever generate one object at a time.
The token is fed to the network through new cross attention layers, which are added to the diffusion UNet and need to be trained.
This is in contrast to other approaches that add conditioning tokens to the existing conditioning via the existing cross attention layers.
The original method uses 2D bounding boxes, described by the $x,y$-coordinates of two opposing corners, \ie 4 channels in total.
We adapt it to our use case by projecting the 8 corners of the 3D bounding box onto the image, and using the $x,y$ coordinates of all of them, meaning 16 channels in total.
While a 3D bounding box can be described fully with only 3 corners, we decided to allow some redundancy by using all 8.

\subsection{Experiment details}
\label{appendix:sec:implementation_details}

\subsubsection{Inputs and conditioning maps}
\label{appendix:sec:model_inputs}
To create the model inputs, we crop the object of interest from a given frame based on the ground truth bounding box centered on the object. During our experiments we noticed that by including more context in the crop our generations looked more realistic, so we further expand the bounding box by a factor of 1.5. In cases where this expansion results in box dimensions exceeding the original image limits, we zero-pad the image and then crop. Finally the input is resized to a spatial dimension of 512x512.

\paragraph{Crop-based inpainting}

At training and inference time, we always operate on square crops around the object instead of the entire frame, for both efficiency and performance reasons. 
For each object, we can derive a 2D bounding box of size $h_b \times w_b$ from the 3D ground truth bounding box corners.
We take a square crop with edges of size $1.5\times\text{max}(h_b, w_b)$, centered on the object, and resize each crop to $512 \times 512$.
Some objects are on the edge of the frame, which would result in aspect ratio issues if resized, and we therefore pad frames to preserve aspect ratio if needed.

\paragraph{3D Bounding box conditioning}

To obtain our proposed pose conditioning map, we start by defining a polygonal mesh template for axis-aligned bounding boxes. Each of the six bounding box faces is made up of two polygons. We assign a color to each, and thus each face is textured with two distinct colors. These consistently indicate which side of the object is visible. Having two colors per face is also helpful when the object is partially out of view: The line separating the two polygons can be used to infer the degree of visibility. Additionally, we represent the eight corners of the bounding box with a spherical mesh. After rendering the box geometry, we overlay a color-coded wireframe on top of the render to make the pose even more explicit, when e.g. only one face of the box is visible.

\paragraph{Depth conditioning.}
To create the depth maps we use for training CN-Depth, we extract the depth maps from the expanded object crops discussed above using Depth-Anything \cite{depthanything}. Similar to the object crop inputs, the depth maps are resized to 512x512.

\paragraph{Loose Control-style conditioning.}

To mimic the conditioning signal of LooseControl, instead of rendering 3D bounding boxes as coloured images indicating pose, we render them to depth maps, as described in \cite{bhat2023loosecontrol} with one notable difference: We do not embed the bounding boxes in a scene depth map to allow for a controlled comparison with our pose-conditioning approach.

\subsubsection{Generator training}
\label{appendix:sec:generator_training}

For models that do not employ additional modules like ControlNets, unless otherwise indicated, we finetune the diffusion UNet for 300,000 steps on the nuScenes training set with a fixed learning rate of $10^{-5}$, which we warm up for 500 steps and set the batch size to 4. We use the AdamW optimizer with $\beta_1 = 0.9$, $\beta_2 = 0.999$, $\epsilon = 10^{-8}$ and a weight decay factor of $0.01$. We use the \texttt{stablediffusion-2-inpainting} checkpoint as the initial state of the UNet weights.

During training we also drop the prompt with probability of 0.5, to enable classifier-free guidance at inference time.
As mentioned in the main text, the prompt we use to guide denoising during training is the simple template \texttt{`an image of a <object class>'}.

\paragraph{ControlNet training.}

In the cases where the model includes a ControlNet, such as our proposed model, ControlNet-Depth and LooseControl, we train the model by jointly finetuning the ControlNet and the upsampling blocks of the UNet using the training setup we discussed above. 
At the start of training, we initialize the weights of the ControlNet using standard procedure: by copying the weights of the UNet, and zero-initializing the connecting convolution layers.

\paragraph{GLIGEN training}

For implementing GLIGEN, we adapted the official implementation\footnote{\url{https://github.com/gligen/GLIGEN}} to our setup. We train it with the same hyper parameters as the other models, \ie 300k steps with batch size 4, for a fair comparison.
The original work uses varying setups for different experiments, but up to 500k steps at batch size 32.

\subsubsection{Exemplar-based conditioning}
\label{appendix:sec:exemplar_conditioning}

\begin{figure*}[t]
\centering
\includegraphics[width=0.8\linewidth]{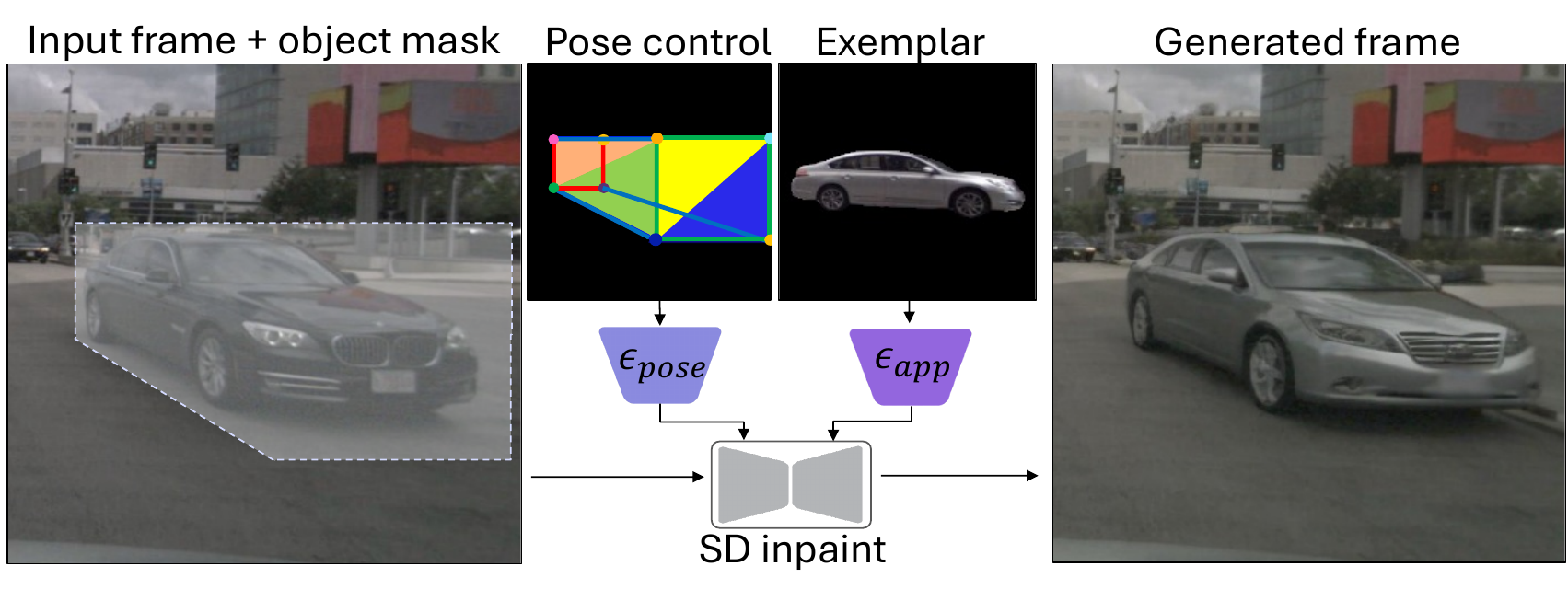}
\caption{Overview of the exemplar-conditioning pipeline. The appearance encoder $\epsilon_{\text{app}}$ creates a conditioning signal from the exemplar image, which shows the object we want to insert.}
\label{fig:exemplar_overview}
\end{figure*}

Exemplar conditioning involves conditioning generation on (a feature representation of) an \emph{exemplar image}, ideally preserving the identity of the exemplar while changing other aspects such as its orientation.
Previous work has shown that a pretrained vision backbone such as CLIP \citep{yang2023paintbyexample} or DINO \cite{caron2021dinov1, wu2024neuralassets} can be effective.
We therefore adopt the same strategy here.
An overview of the architecture is shown in Fig.~\ref{fig:exemplar_overview}.

We initialize a DINO v1 backbone and freeze its weights. 
We then add a transformer output head, consisting of two transformer blocks that see all $28 \times 28 + 1$ DINO output tokens.
Each transformer block is a \texttt{BasicTransformerBlock} as implemented in the diffusers library with dimensionality 768, no dropout, and using a single attention head.
We take the first $4$ tokens in the output sequence as the conditioning for generation, and finetune the diffusion UNet and trainable layers of the exemplar encoder jointly.
The choice to use 4 tokens instead of 1 is mainly because we observed better exemplar fidelity for the amount of training we used here.
Together, the DINO backbone and this custom output head form our appearance encoder $\epsilon_\text{app}$.

Getting the exemplar from the same frame as the target frame is potentially suboptimal, as the model may learn to simply copy the exemplar into the frame.
During training, we therefore get the exemplars from adjacent frames.
Following \cite{yang2023paintbyexample}, we also augment the exemplar image with light augmentation, applying horizontal flipping with probability 0.5, and randomly rotating the exemplar by up to 20 degrees uniformly at random.
At inference time, we always flip the exemplar horizontally and use no rotation.

\subsection{Additional evaluations}
\label{appendix:sec:additional_evaluations}

\setlength{\tabcolsep}{3pt} 
\begin{table*}[t!]
\footnotesize{
\caption{Object editing performance in the replacement task. The bottom left table contains the full evaluation results on the standard replacement setting. The bottom right table contains the full evaluation results on the replacement with flipped pose instruction setting. All results were obtained by averaging per object class average metrics. Oracle results are obtained on ground truth data. FID is computed with respect to a hold-out set of 5,000 objects.}
\def\arraystretch{1.2}
    \begin{minipage} {.48\textwidth}
        \centering
        \label{tab:app_object_editing}
        \begin{tabular}{lrrrrrr}
            \toprule
            Method            &  mAOE$\downarrow$ & Flips$\downarrow$ & mATE$\downarrow$  & Conf.$\uparrow$ & CLIP$\uparrow$ & FID$\downarrow$ \\
            \midrule                   
            Oracle                                       & 0.02 & 0.00 & 0.35 & 0.70 & 21.89 & 5.60 \\
            \midrule                   
            SD-Inpaint \cite{rombach2022stablediffusion} & 0.76 & 0.24 & 1.40 & 0.64 & 22.05 & 10.47 \\
            LooseControl \citep{bhat2023loosecontrol}    & 0.87 & 0.27 & 1.54 & 0.64 & \textbf{22.88} & 14.37 \\
            GLIGEN \citep{gligen}                        & 0.26 & 0.04 & 1.61 & 0.59 & 21.67 & 16.99 \\
            GLIGEN (500k) \citep{gligen}                 & 0.20 & 0.03 & 1.43 & 0.62 & 21.76 & 16.85 \\
            Neural Assets                                & 0.26 & 0.04 & 1.55 & 0.60 & 21.74 & 14.69 \\
            Neural Assets (400k)                         & 0.20 & 0.03 & 1.54 & 0.61 & 21.90 & 11.98 \\
            CN-Depth \citep{controlnet}                  & 0.13 & 0.02 & \textbf{0.784} & \textbf{0.67} & 21.90 & \textbf{7.98} \\
            Ours                                         & \textbf{0.12} & \textbf{0.01} & 1.39 & 0.65 & 22.05 & 9.37 \\
            \bottomrule
        \end{tabular}
    \end{minipage}
    \hfill
    \begin{minipage}{.48\textwidth}
        \centering
        \begin{tabular}{lrrrrrr}
            \toprule
            Method            &  mAOE$\downarrow$ & Flips$\downarrow$ & mATE$\downarrow$ & Conf.$\uparrow$ & CLIP$\uparrow$ & FID$\downarrow$  \\
            \midrule        
            SD-Inpaint    \cite{rombach2022stablediffusion} & 2.38  & 0.24 & 1.40 & 0.64 & 22.05 & 10.47 \\
            LooseControl  \citep{bhat2023loosecontrol}      & 2.29  & 0.27 & 1.54 & 0.64 & \textbf{22.80} & 14.37 \\
            GLIGEN        \citep{gligen}                    & 0.65  & 0.14 & 1.79 & 0.56 & 21.70 & 17.55 \\
            GLIGEN (500k)        \citep{gligen}             & 0.49  & 0.09 & 1.61 & 0.58 & 21.80 & 17.21 \\
            Neural Assets                                   & 0.50  & 0.11 & 1.71 & 0.57 & 21.79 & 15.81 \\
            Neural Assets (400k)                            & 0.39  & 0.08 & 1.71 & 0.58 & 21.91 & 13.07 \\
            CN-Depth      \citep{controlnet}                & 3.01  & 0.97 & \textbf{0.78} & \textbf{0.67} & 21.90 & \textbf{7.98} \\
            Ours           & \textbf{0.36}                  & \textbf{0.07} & 1.507 & 0.63 & 22.13 & 9.78 \\
            \bottomrule
        \end{tabular}
    \end{minipage}
}

\end{table*}

Table~\ref{tab:app_object_editing} contains the complete evaluation results for all models in the replacement task, including the aggregate confidence score of the pretrained Epro-PnP detector. We include this score as an additional metric to quantify difference in model performance with respect to the success of the edit, given the fact that the average CLIP scores for the evaluated models exhibit very small variance. 
Confidence scores seem to exhibit the same behaviour, meaning that all models edit a given object with comparable quality.

With the exception of CLIP and FID, the results shown in Table~\ref{tab:app_object_editing} are obtained by averaging the class averages of all metrics. In Table~\ref{tab:app_class_ablation_aoe} we show the per-class performance of all models with respect to all relevant metrics. As also shown in table~\ref{tab:class_ablation_aoe} in the main paper, our approach performs robustly, even in rare class cases. We add the full table here for completeness. We note that \emph{Flips} denotes the ratio of the edited objects with flipped orientation over the total number of objects in the specific class. 
We consider the orientation of an edited object flipped when its orientation error with respect to the original object exceeds 90 degrees or 1.57 radians.

\begin{table*}[t!]
\centering

\footnotesize{
    \caption{Per-class metrics on the object replacement task. Below, the Average Orientation Error (in radians), Average Translation Error (in meters), the percentage of orientation flips and the 3D detector confidence score are shown for each vehicle class in our validation set.}
    \label{tab:app_class_ablation_aoe}
    \centering
    \def\arraystretch{1.2}

    \begin{tabular}{lrrrrrrrrrrrr} 
        \toprule 
        Category    & \multicolumn{4}{c}{cars} & \multicolumn{4}{c}{trucks} & \multicolumn{4}{c}{buses} \\ 
        \# instances & \multicolumn{4}{c}{3871} & \multicolumn{4}{c}{905}    & \multicolumn{4}{c}{224}  \\ 
        \midrule
                        & AOE$\downarrow$           & ATE$\downarrow$           & Flips$\downarrow$ & Conf$\uparrow$ & AOE$\downarrow$ & ATE$\downarrow$           & Flips$\downarrow$ & Conf$\uparrow$ & AOE$\downarrow$ & ATE$\downarrow$           & Flips$\downarrow$ & Conf$\uparrow$ \\
        \hline
        SD-Inpaint      & 0.76          & 2.98          & 0.22 & 0.73 & 0.97 & 4.76 & 0.28 & 0.54 & 0.65 & 6.02 & 0.18 & 0.65 \\
        LooseControl    & 0.96          & 3.09          & 0.29 & 0.72 & 1.00 & 4.89 & 0.30 & 0.55 & 0.83 & 6.19 & 0.24 & 0.64 \\
        GLIGEN          & 0.22          & 3.00          & 0.03 & 0.71 & 0.39 & 4.64 & 0.08 & 0.50 & 0.34 & 6.97 & 0.07 & 0.58 \\ 
        GLIGEN (500k)          & 0.18          & 2.96          & 0.02 & 0.72 & 0.36 & 4.41 & 0.07 & 0.52 & 0.24 & 6.04 & 0.04 & 0.62 \\ 
        Neural Assets          & 0.20          & 3.03          & 0.03 & 0.69 & 0.35 & 4.53 & 0.07 & 0.50 & 0.40 & 5.95 & 0.10 & 0.60 \\ 
        Neural Assets (400k)         & 0.18          & 3.04          & 0.02 & 0.71 & 0.31 & 4.76 & 0.06 & 0.50 & 0.27 & 6.32 & 0.06 & 0.63 \\ 
        CN-Depth        & \textbf{0.15} & \textbf{2.87} & \textbf{0.01} & \textbf{0.75} & 0.21 & \textbf{4.22} & \textbf{0.02} & \textbf{0.58} & 0.24 & 5.79 & 0.04 & \textbf{0.68} \\ 
        Ours            & \textbf{0.15} & 2.97          & \textbf{0.01} & 0.74 & \textbf{0.19} & 4.64 & \textbf{0.02} & 0.56 & \textbf{0.16} & \textbf{5.78} & \textbf{0.01} & 0.65 \\ 
        \bottomrule 
    \end{tabular}
}

\end{table*}

\subsection{Additional qualitative results}
\label{appendix:sec:additional_visuals}
\begin{figure*}[t]
\centering
\includegraphics[width=\linewidth]{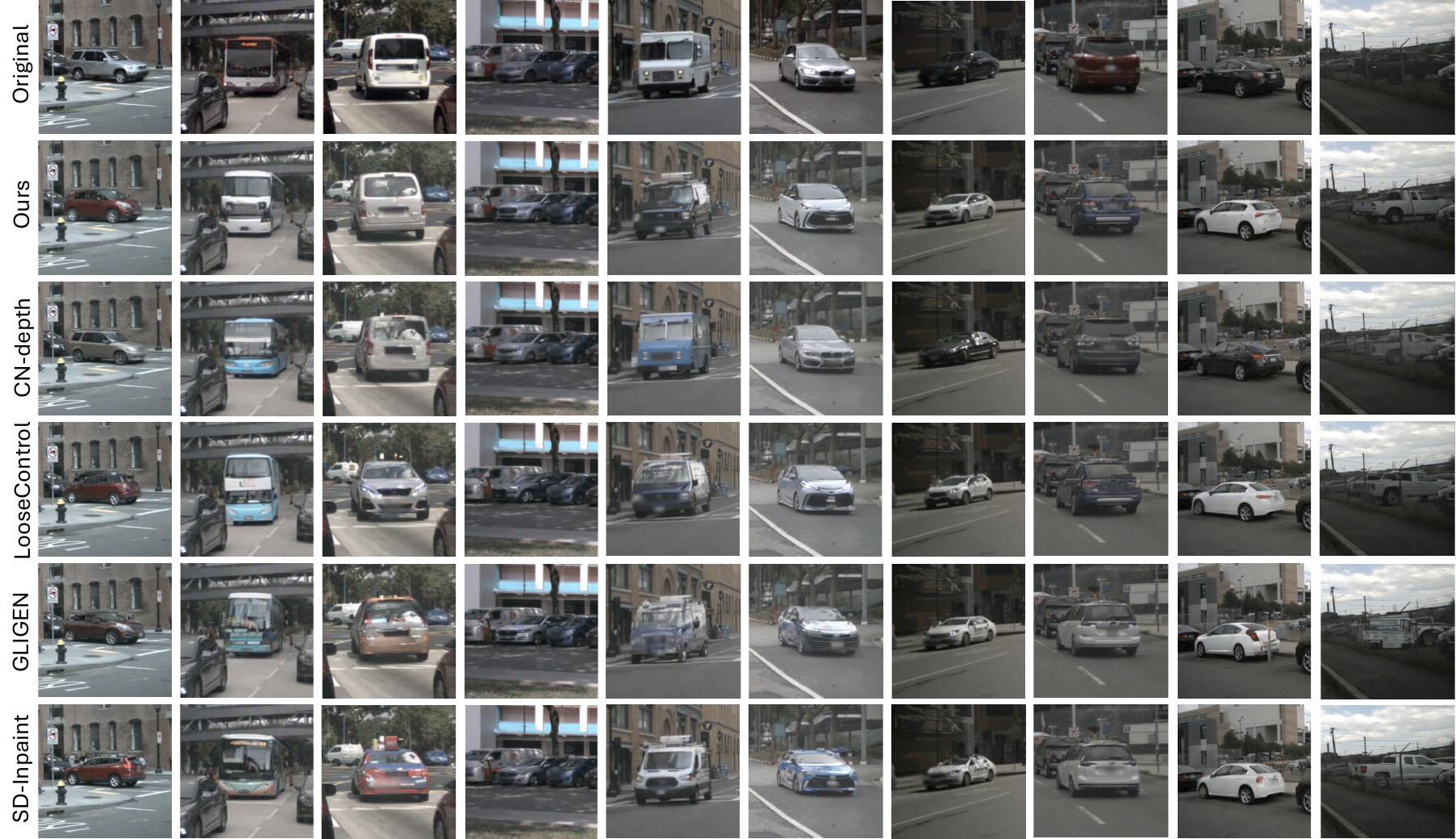}
\caption{
Example generations from all models in the standard object editing evaluation setting.
}
\label{fig:app_replacement_samples}
\end{figure*}

\begin{figure*}[t]
\centering
\includegraphics[width=\linewidth]{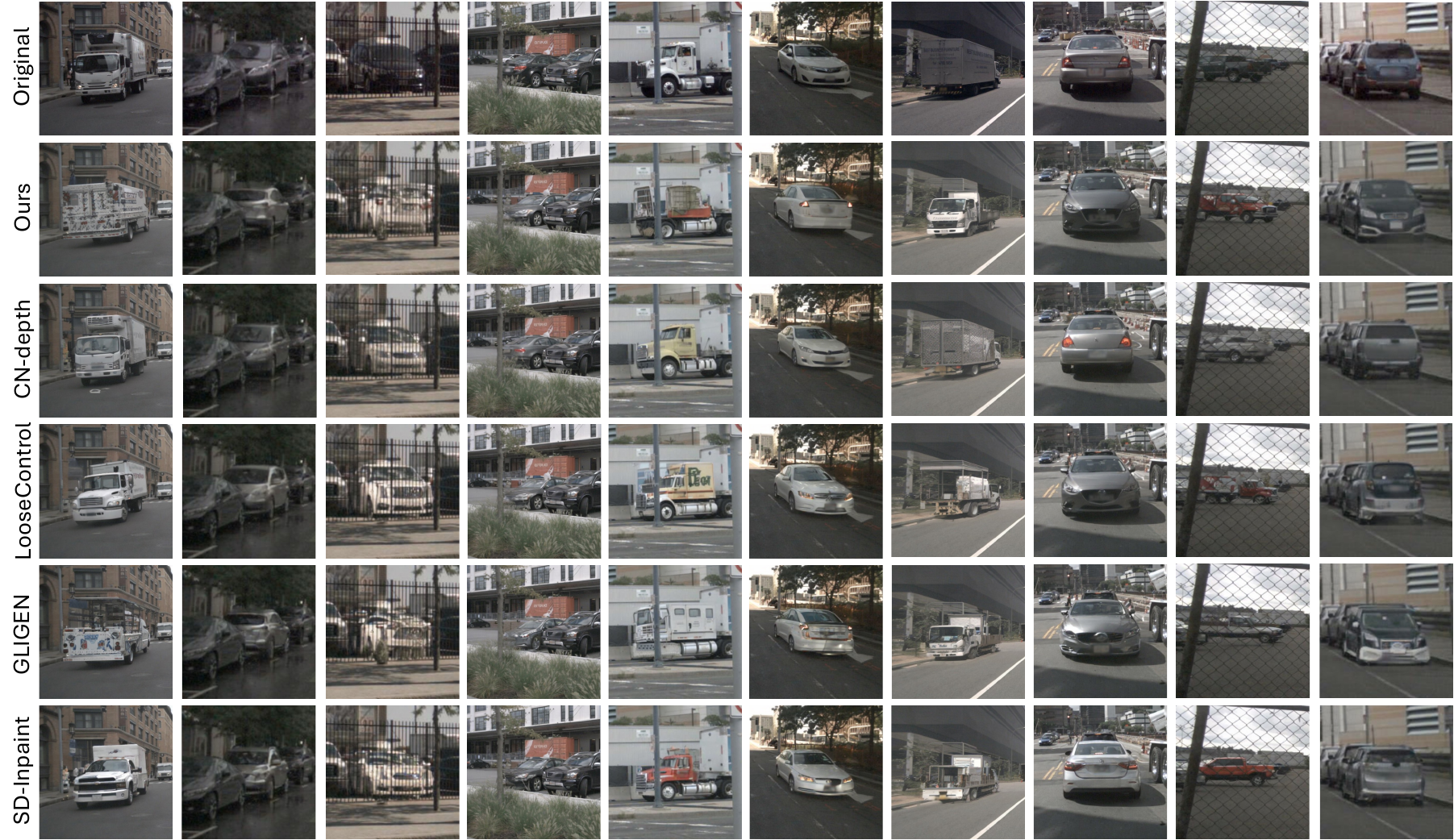}
\caption{
Example generations from all models in the object editing evaluation setting with flipped pose instruction.
}
\label{fig:app_replacement_samples_flipped}
\end{figure*}
Figures~\ref{fig:app_replacement_samples} and~\ref{fig:app_replacement_samples_flipped} show randomly sampled edited objects from the evaluation set for all models for the replacement and replacement with flipped orientation instruction tasks respectively. Our conditioning and GLIGEN are the only ones flexible enough to consistently apply the desired edit, as opposed to the rest of the evaluated baselines. Concordantly, ours and GLIGEN are the only models which avoid large orientation errors (i.e. a ``flip'' of the intended orientation), with only 10\% of all evaluation cases exhibiting such errors in our case. Furthermore, our model has the best mAOE performance among all evaluated models.

\begin{figure*}
    \centering
    \includegraphics[width=0.97\linewidth]{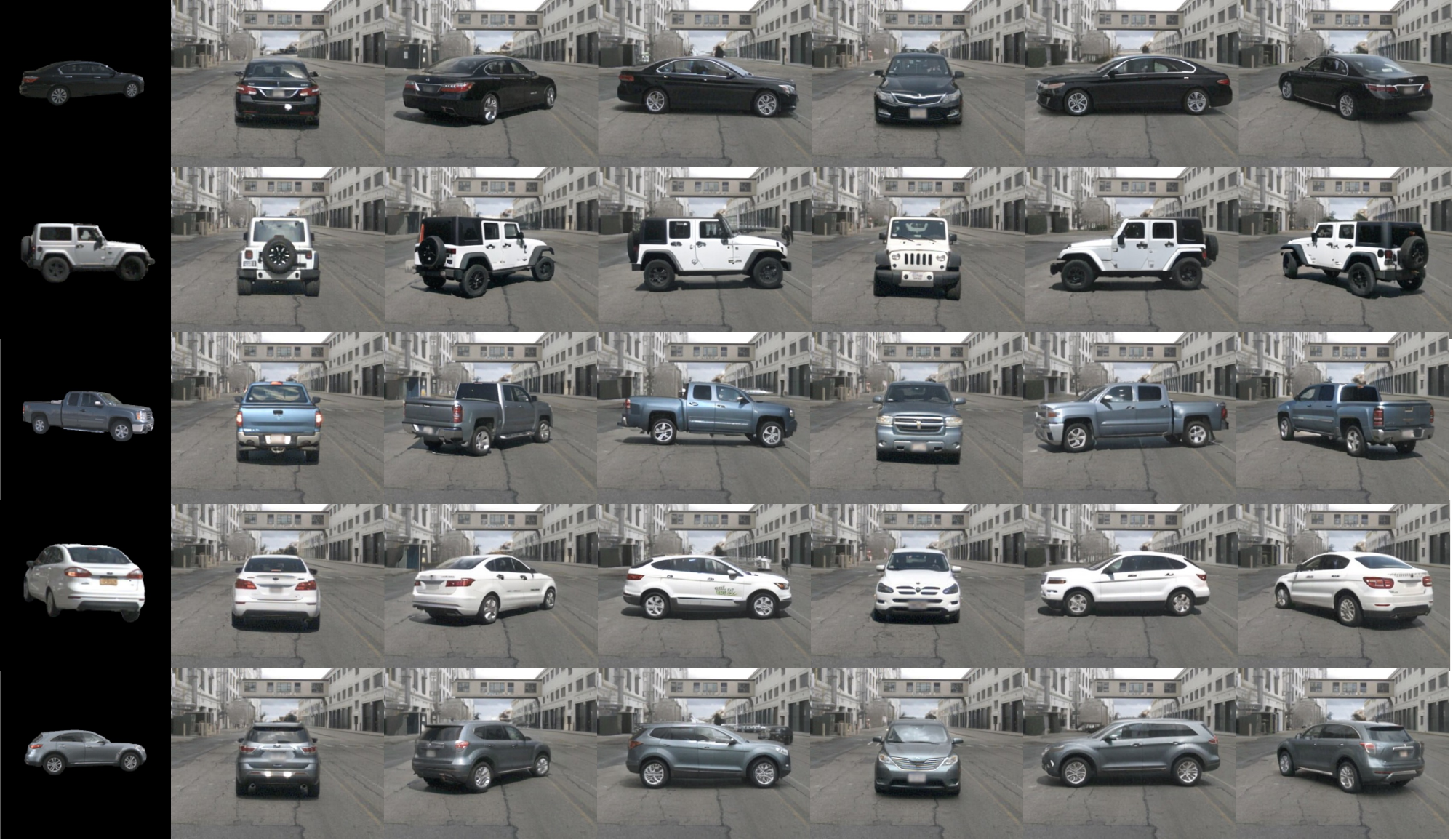}
    \caption{Placing specific vehicles in specific locations. The left column shows the exemplar, the other columns show a the result of generation with a placement instruction. 
    We show a $400 \times 500$ crop of the $900 \times 1600$ image.
    }
    \label{fig:appendix:exemplar_placement_seq0}
\end{figure*}
\begin{figure*}
    \centering
    \includegraphics[width=0.97\linewidth]{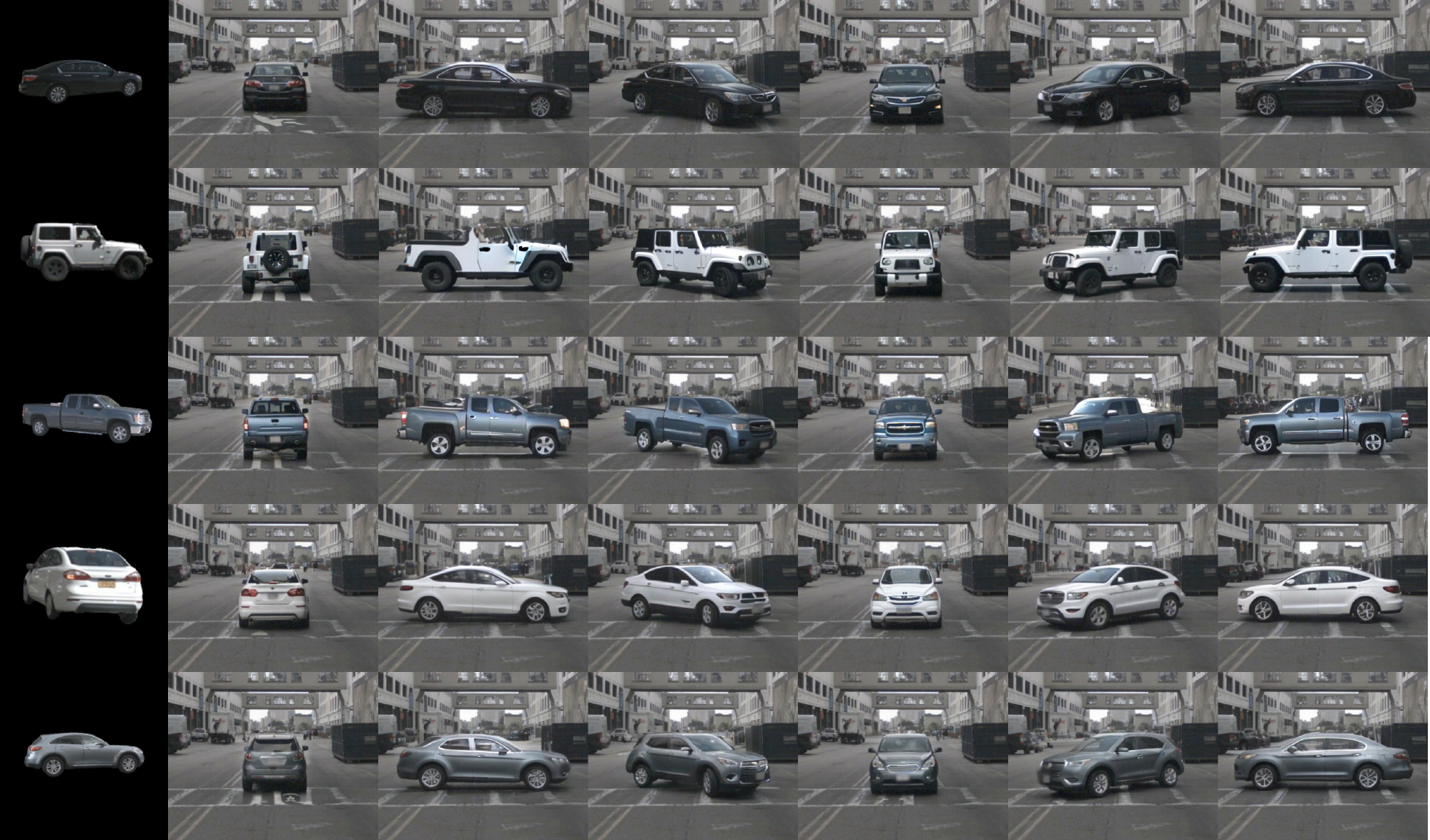}
    \caption{Placing specific vehicles in specific locations. The left column shows the exemplar, the other columns show a the result of generation with a placement instruction.
    We show a $400 \times 500$ crop of the $900 \times 1600$ image.
    }
    \label{fig:appendix:exemplar_placement_seq5}
\end{figure*}
\begin{figure*}
    \centering
    \includegraphics[width=0.97\linewidth]{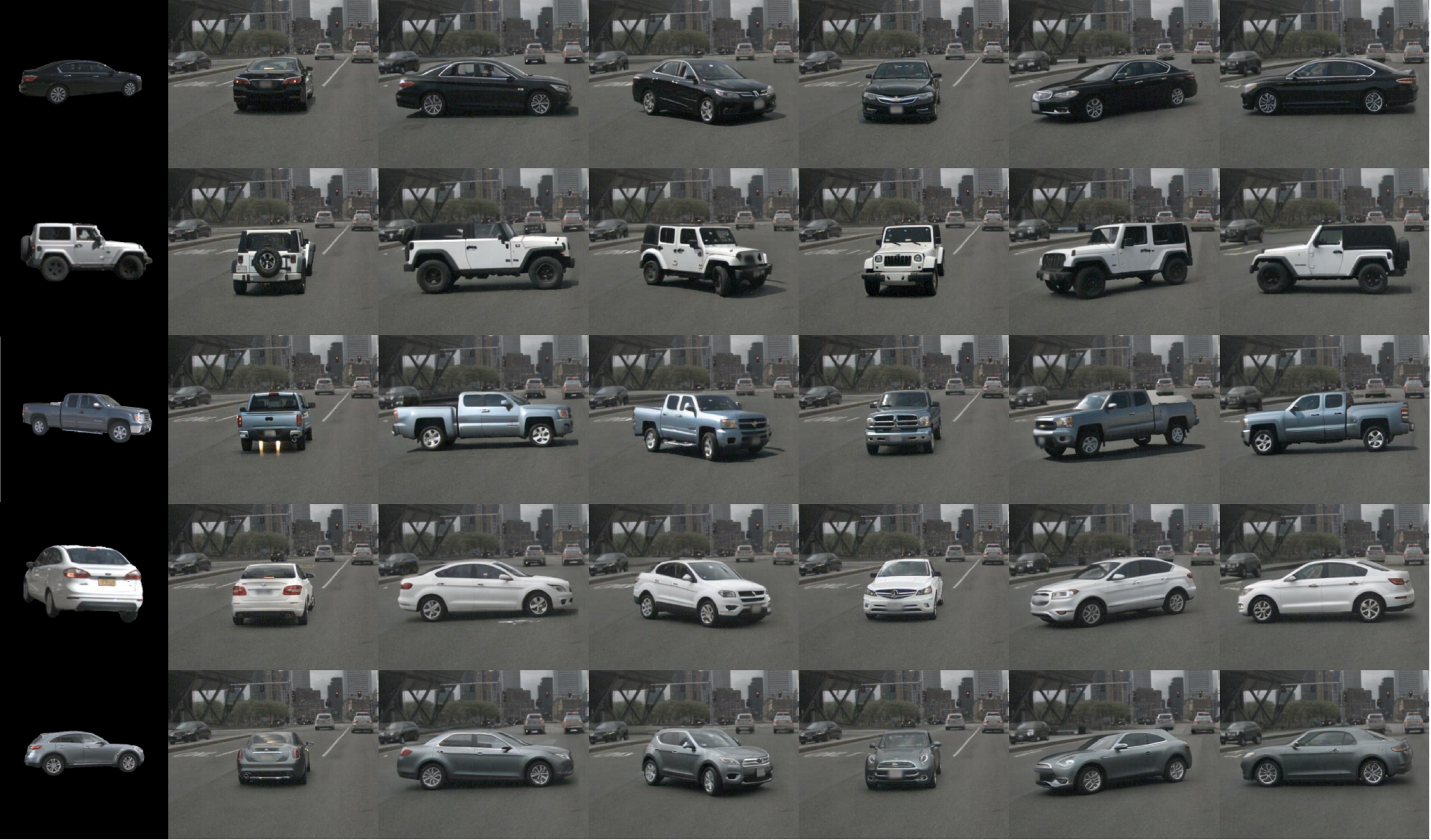}
    \caption{Placing specific vehicles in specific locations. The left column shows the exemplar, the other columns show a the result of generation with a placement instruction.
    We show a $400 \times 500$ crop of the $900 \times 1600$ image.
    }
    \label{fig:appendix:exemplar_placement_seq6}
\end{figure*}

We show additional results for exemplar-based placement in Figures \ref{fig:appendix:exemplar_placement_seq0}, \ref{fig:appendix:exemplar_placement_seq5} and \ref{fig:appendix:exemplar_placement_seq6}.
For each of these figures, we place our box conditioning in six locations in scenes obtained from the nuScenes front camera.
This shows that we are able to preserve the identity of the exemplar, even when this exemplar is facing a different direction than the desired orientation.

\end{document}